\documentclass{article}

\usepackage[preprint]{neurips_2026}

\usepackage[utf8]{inputenc} 
\usepackage[T1]{fontenc}    
\usepackage{hyperref}       
\usepackage{url}            
\usepackage{booktabs}       
\usepackage{amsfonts}       
\usepackage{nicefrac}       
\usepackage{microtype}      
\usepackage{xcolor}         
\usepackage{amsmath}
\usepackage{graphicx}                         
\usepackage{amssymb}       
\usepackage{algorithm}
\usepackage{algorithmic}
\usepackage{enumitem}       
\usepackage{multirow}
\usepackage{subcaption}
\usepackage{colortbl}
\usepackage{wrapfig}
\usepackage{listings}
\definecolor{codegreen}{rgb}{0.0,0.5,0.0}
\definecolor{codegray}{rgb}{0.5,0.5,0.5}
\definecolor{codeblue}{rgb}{0.0,0.0,0.7}
\definecolor{codepurple}{rgb}{0.58,0,0.82}
\definecolor{backcolour}{rgb}{0.97,0.97,0.97}
\lstdefinestyle{pythonstyle}{
    backgroundcolor=\color{backcolour},
    commentstyle=\color{codegreen},
    keywordstyle=\color{codeblue}\bfseries,
    stringstyle=\color{codepurple},
    basicstyle=\ttfamily\scriptsize,
    breaklines=true,
    numbers=left,
    numberstyle=\tiny\color{codegray},
    numbersep=5pt,
    xleftmargin=12pt,
    framexleftmargin=12pt,
    language=Python,
    morekeywords={Tensor, list, tuple, Linear, RMSNorm},
    showstringspaces=false,
}

\usepackage{amsthm}

\usepackage{pifont}
\usepackage{arydshln}

\newcommand{\cmark}{\ding{51}}  
\newcommand{\xmark}{\ding{55}}
\title{xHC: Expanded Hyper-Connections}

\author{
  \textbf{Xiangdong Zhang}$^{1,2}$ \quad
  \textbf{Xiaohan Qin}$^{1,2}$ \quad
  \textbf{Sunan Zou}$^{2,4}$ \quad
  \textbf{Tuo Dai}$^{2}$ \quad
  \textbf{Xiaoming Shi}$^{2}$ \\[2pt]
  \textbf{Huaijin Wu}$^{1,2}$ \quad
  \textbf{Yebin Yang}$^{1,2}$ \quad
  \textbf{Zhuo Xia}$^{1,2}$ \quad
  \textbf{Shaofeng Zhang}$^{3}$ \quad
  \textbf{Lin Yao}$^{2}$ \\[2pt]
  \textbf{Yuliang Liu}$^{5}$ \quad
  \textbf{Yu Cheng}$^{6}$ \quad
  \textbf{Junchi Yan}$^{1}$\thanks{Corresponding author: \texttt{yanjunchi@sjtu.edu.cn}. Project page: \url{https://github.com/aHapBean/xHC}.} \\[8pt]
  {\normalfont $^1$School of AI, Shanghai Jiao Tong University \quad
  $^2$Dots Studio, Xiaohongshu Inc.} \\
  {\normalfont $^3$University of Science and Technology of China \quad
  $^4$School of CS, Peking University} \\
  {\normalfont $^5$Independent Researcher \quad
  $^6$The Chinese University of Hong Kong}
}

\begin{document}

\maketitle
\vspace{-15pt}
\begin{abstract}
\vspace{-5pt}

Hyper-Connections (HC) expand the residual stream of Transformers into $N$ parallel streams, providing a form of memory scaling beyond model width and depth. Manifold-Constrained HC (mHC) stabilizes this formulation at scale. The large gains from $N{=}1$ to $N{=}4$ suggest residual-stream expansion as a promising scaling axis. However, existing HC-family methods typically stop at $N{=}4$. Our experiments reveal why: scaling mHC beyond this point yields diminishing performance gains and rapidly increasing training cost. We attribute this limitation to two bottlenecks: insufficient write-back information for an expanding number of streams and residual-mixing generation whose cost scales cubically with $N$.
To address both bottlenecks, we propose \textbf{xHC} (E\textbf{x}panded \textbf{H}yper-\textbf{C}onnections), the first HC-family method to achieve meaningful expansion beyond $N{=}4$. xHC combines temporal feature augmentation for richer write-back with a sparse residual-stream architecture that updates only $k=4$ of the $N=16$ streams while retaining dense access to the full residual state. \textbf{Across 18B and 28B MoE models, xHC delivers strong and consistent downstream improvements.} On an 18B MoE model, xHC improves the average downstream score by 4.0 points over mHC, while adding only modest training FLOPs over the vanilla baseline. Scaling-law experiments show that the vanilla and mHC baselines require $1.50\times$ and $1.19\times$ the compute of xHC, respectively, to reach the same loss. 
Practical large-$N$ training also requires controlling memory traffic from the expanded residual state. We therefore introduce \textbf{xHC-Flash}, which reduces the per-sublayer memory traffic from $73.5C$ to $40C$, comparable to the $34C$ required by mHC at $N{=}4$, while retaining most of the gains of full xHC. 
Together, xHC and xHC-Flash make large-$N$ residual-stream expansion effective and practical for LLM pre-training.

\end{abstract}

\vspace{-13pt}
\section{Introduction}
\vspace{-2pt}

\begin{wrapfigure}{r}{0.48\textwidth}
    \vspace{-8pt}
    \centering
    \vspace{-0.8em}
    \includegraphics[width=\linewidth]{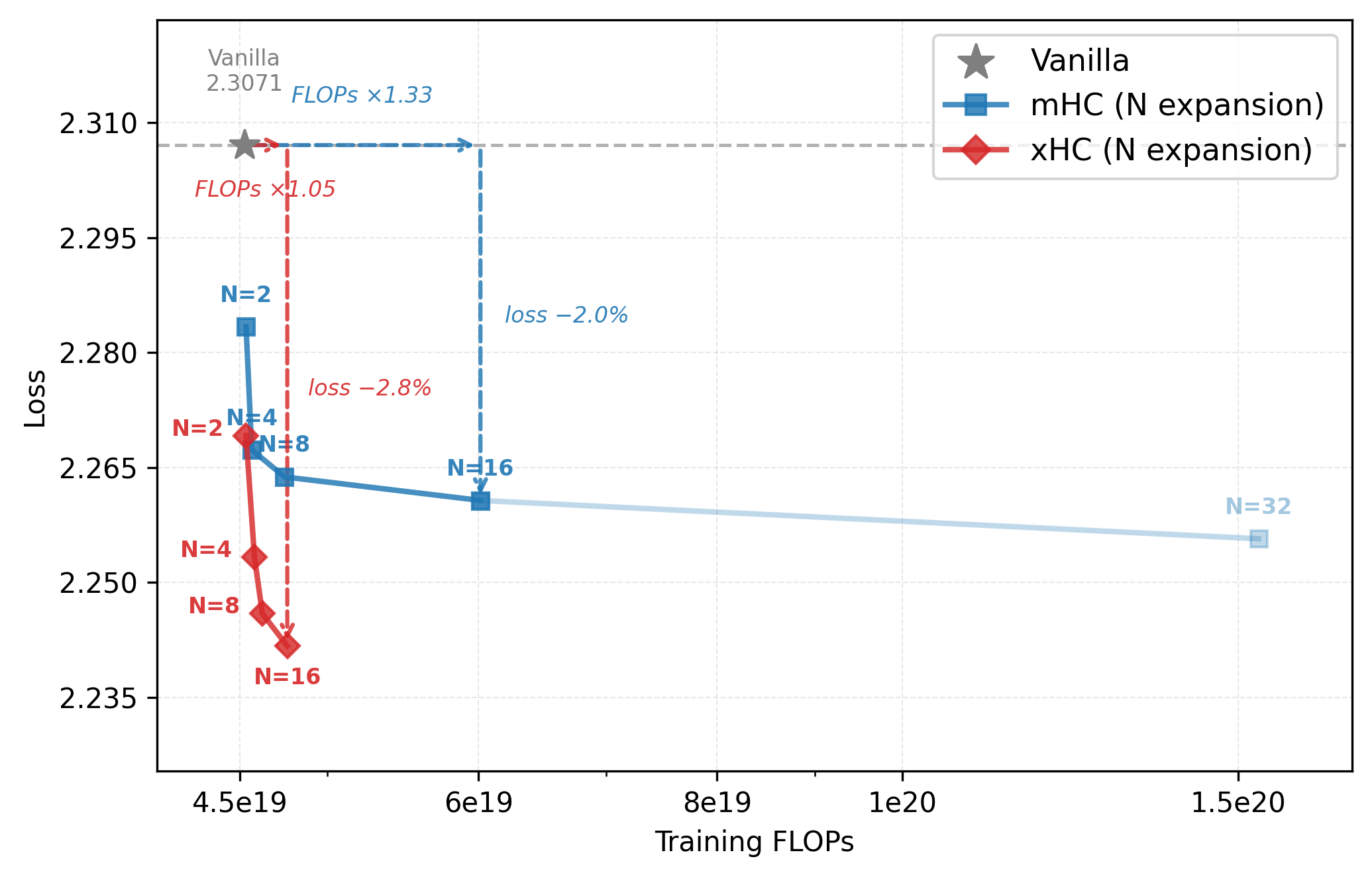}

    \vspace{-7pt}
    \caption{Expansion efficiency: loss vs.\ FLOPs on a 2.5B MoE model (details in Table~\ref{tab:hyperparams}).}
    \label{fig:teaser_expansion}
    \vspace{-1.0em}
\end{wrapfigure}

Large language model architectures have advanced rapidly~\citep{liu2024deepseek,yang2025qwen3,grattafiori2024llama,liu2025deepseek32,team2025kimi2,yang2026jtok} through improved attention mechanisms~\citep{vaswani2017attention,NSA,shazeer2019MQA,ainslie2023gqa}, Mixture-of-Experts~\citep{dai2024deepseekmoe,jiang2024mixtral}, and scaling in width, depth, and data~\citep{kaplan2020scaling,hoffmann2022chinchilla}.
Yet the residual stream that carries token representations across layers remains largely unchanged: it still operates as a single identity pathway, offering no learnable control over cross-layer information flow~\citep{he2016deep}.
Hyper-Connections (HC)~\citep{zhu2024hyper} challenge this design by replacing the single stream with $N$ parallel residual streams governed by learnable mixing matrices.
Building on HC, Manifold-Constrained HC (mHC)~\citep{DS_mHC} makes multi-stream residual training stable at scale and represents the state-of-the-art variant.
Residual-stream expansion increases the persistent state of models with modest additional FLOPs, offering a form of \emph{memory scaling} beyond width and depth. The strong gains observed when expanding from $N{=}1$ to $N{=}4$ in prior results~\citep{zhu2024hyper} suggest that this is a promising scaling axis.
However, existing HC-family methods~\citep{zhu2024hyper,DS_mHC,yang2026mhclite,sHC} typically stop at $N{=}4$.
We study this through mHC, the state-of-the-art HC formulation at scale.
Our experiments in Figure~\ref{fig:teaser_expansion} show that naively increasing $N$ further yields rapidly diminishing returns: loss improves only marginally from $N{=}4$ to $N{=}16$, while training FLOPs increase by 32\%, leaving this scaling axis largely unrealized.

We argue that this benefit-cost imbalance is not incidental, but stems from two distinct bottlenecks: one limits the benefit of larger $N$, while the other increases its cost.
The first is an \textbf{information bottleneck}: each stream is meant to store a different weighted history of layer outputs, but each layer injects only a single write-back signal into the multi-stream state.
As $N$ grows, forming meaningful and diverse stream histories requires richer write-back information than this single signal can provide, making additional streams increasingly redundant.
The second is a \textbf{computational bottleneck}: in mHC, the dominant cost comes from generating the residual mapping $\mathcal{H}_\text{res} \in \mathbb{R}^{N \times N}$ through an input-dependent projection.
Because this projection predicts $N^2$ coefficients from an $NC$-dimensional residual state, its cost scales as $O(N^3 C)$, making the cost of expansion grow much faster than its performance benefit.
Together, these bottlenecks explain why simply adding more streams does not translate into proportionally more useful capacity.
They also clarify what it would take to make expansion rate a true scaling axis: the model must supply more diverse write-back information while avoiding the cubic cost of scaling $\mathcal{H}_\text{res}$.

\begin{figure*}[t]
    \centering
    \begin{subfigure}[t]{0.325\textwidth}
        \centering
        \includegraphics[width=\linewidth]{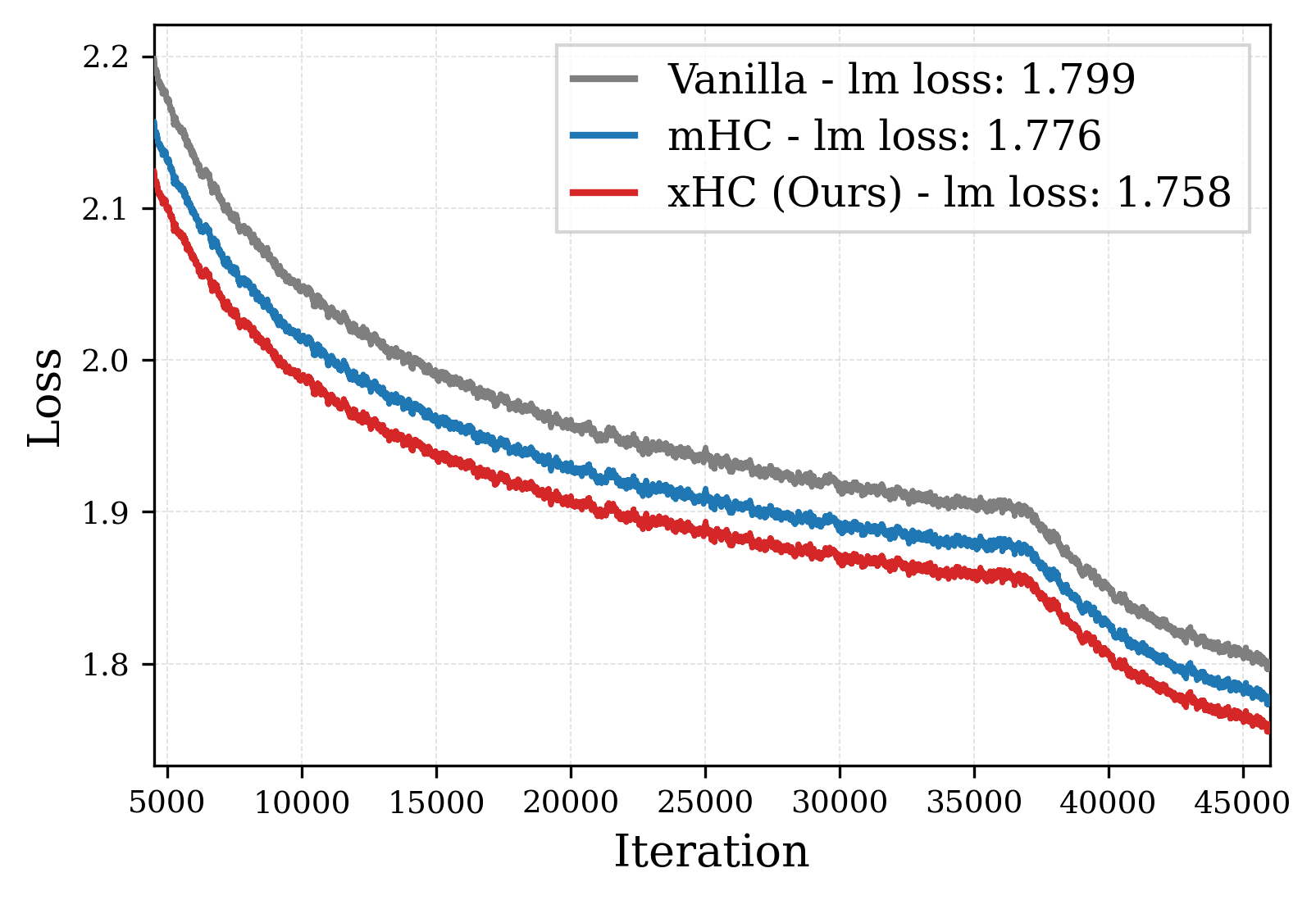}
        \caption{Training loss.}
        \label{fig:teaser_loss}
    \end{subfigure}
    \hfill
    \begin{subfigure}[t]{0.325\textwidth}
        \centering
        \includegraphics[width=\linewidth]{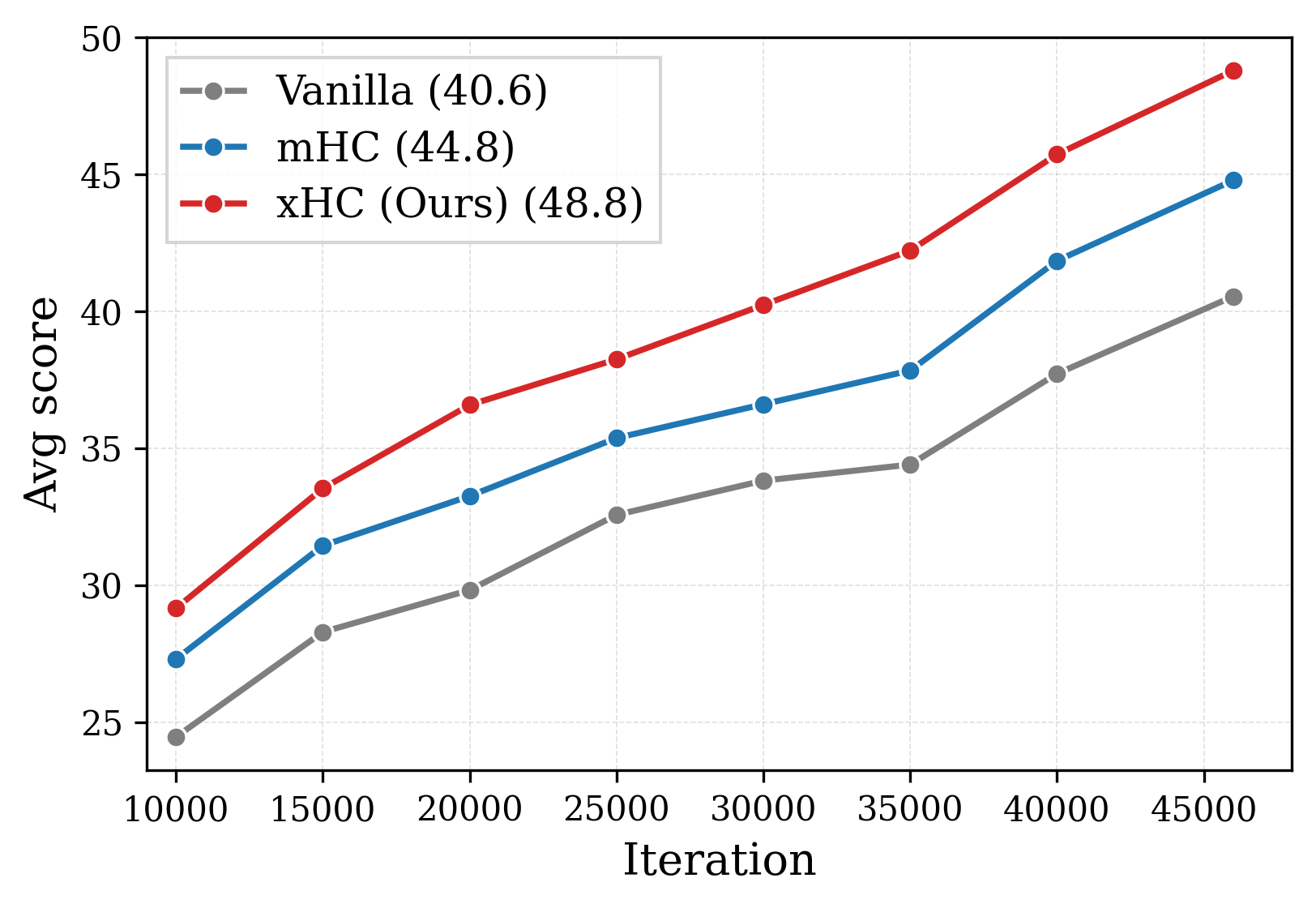}
        \caption{Average downstream score.}
        \label{fig:teaser_perf}
    \end{subfigure}
    \hfill
    \begin{subfigure}[t]{0.325\textwidth}
        \centering
        \includegraphics[width=\linewidth]{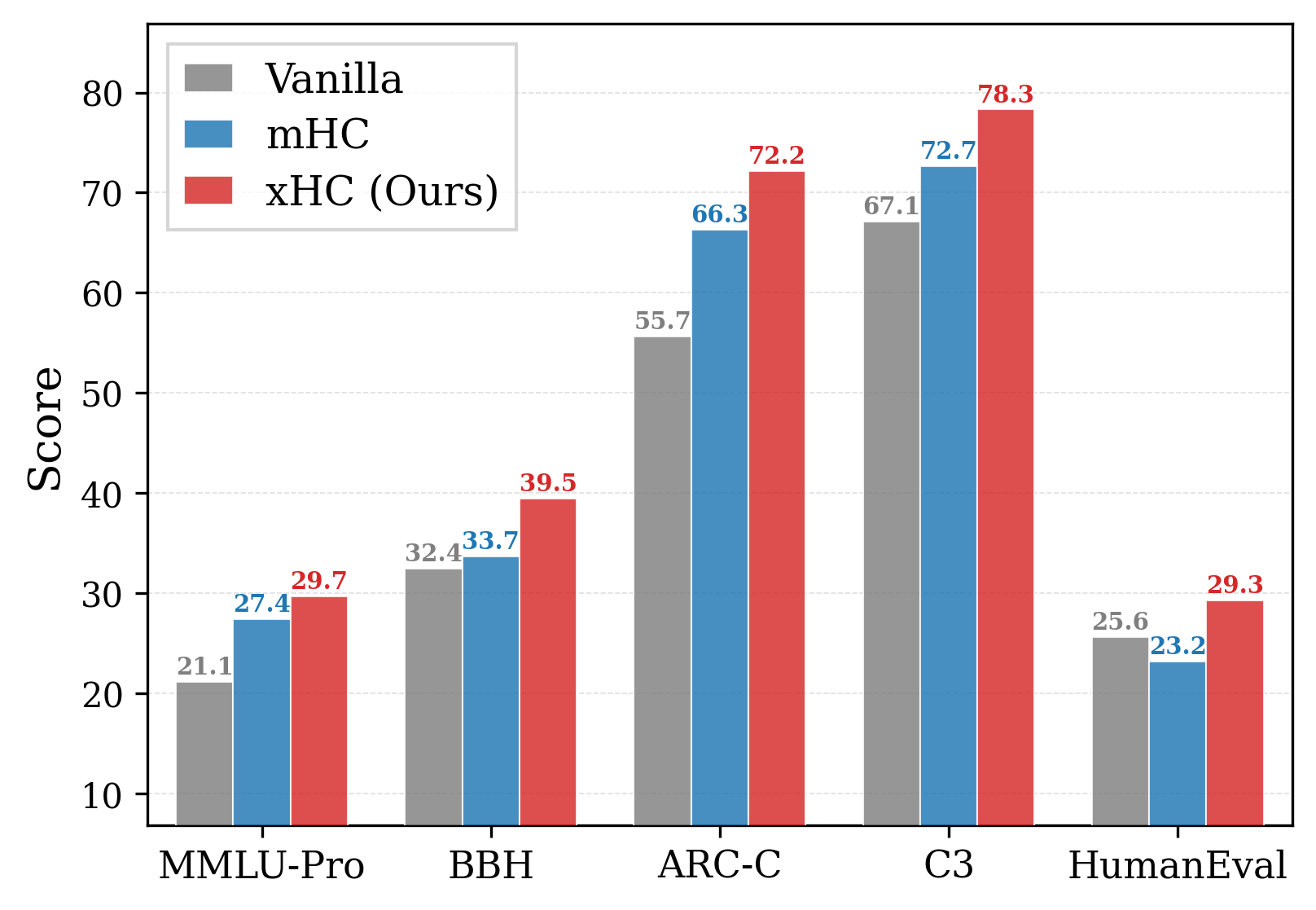}
        \caption{Benchmark-level gains.}
        \label{fig:teaser_benchmarks}
    \end{subfigure}
    \caption{\textbf{xHC delivers broad gains at 18B scale.} (a) xHC achieves lowest training loss. (b) The loss improvement translates into a significantly higher average downstream score across benchmarks in Table~\ref{tab:main_results}. (c) xHC improves representative benchmarks across reasoning, knowledge, and code.}
    \label{fig:teaser}
    \vspace{-20pt}
\end{figure*}

We therefore propose \textbf{xHC} (E\textbf{x}panded \textbf{H}yper-\textbf{C}onnections), the first method to achieve \emph{meaningful} expansion of HC-family models beyond $N{=}4$.
Building on the stable multi-stream formulation of mHC, xHC makes large-$N$ expansion both effective and affordable through two complementary designs.
To increase the benefit of larger $N$, xHC introduces {temporal feature augmentation} along the causal token sequence: it enriches the write-back signal with low-cost features from neighboring tokens, using multi-scale causal convolutions to supply more diverse information as the number of streams grows.
To control the cost of larger $N$, xHC introduces a sparse residual-stream architecture that activates only $k$ out of $N$ streams for residual mixing and write-back, while keeping the read path dense so that every layer can still access the full $N$-stream state.
These two designs are structurally decoupled yet highly synergistic: as $N$ grows, temporal augmentation becomes more useful because additional streams require richer write-back information, while sparse residual updates become more valuable because they substantially reduce the extra FLOPs introduced by large-$N$ expansion.

Empirically, xHC delivers substantial gains over both mHC and the vanilla residual baseline.
As shown in Figure~\ref{fig:teaser}, xHC consistently improves both pre-training loss and downstream performance at the 18B MoE scale.
It reaches a lower final training loss than mHC and the vanilla baseline (1.758 vs.\ 1.776 and 1.799), and improves the average downstream score from 44.8 with mHC to 48.8, while adding only 4.1\% training FLOPs relative to the vanilla baseline.
xHC also remains effective when the backbone is trained with Muon, indicating that its gains are not specific to AdamW.

More importantly, xHC changes the benefit-cost tradeoff of scaling the expansion rate itself.
As shown in Figure~\ref{fig:teaser_expansion}, on a 2.5B MoE model, increasing $N$ from 4 to 16 in mHC reduces loss by only 0.006 while increasing training FLOPs by 32\%.
In contrast, increasing $N$ from 4 to 16 in xHC reduces loss by 0.012 with only 4\% extra FLOPs.
This shows that xHC makes larger $N$ substantially more cost-effective, turning residual-stream expansion into a practical scaling axis.

This improved benefit-cost tradeoff also translates into better compute efficiency across model scales.
Our scaling-law experiments in Figure~\ref{fig:scaling_law} show that, to reach the same loss, the vanilla requires $1.50\times$ compute of xHC and mHC requires $1.19\times$ compute of xHC.
This confirms that the improvement is systematic rather than confined to a specific model size.

Together, these results establish that xHC improves both the utility and FLOPs efficiency of residual-stream expansion. Practical large-$N$ training, however, also requires controlling the memory traffic introduced by the expanded residual state. We therefore introduce xHC-Flash, which shares routing and dense read computation across sublayers to avoid repeated full-state accesses. It reduces the estimated per-sublayer traffic from $73.5C$ to $40C$, comparable to the $34C$ required by mHC at $N{=}4$, while retaining nearly all of the performance gains of full xHC. We further develop a fused implementation that consolidates residual-stream operations, reducing memory traffic and kernel launch overhead. \textbf{Our contributions are as follows:}

\textbf{1) Bottleneck diagnosis.} We identify two bottlenecks behind the diminishing returns of larger $N$ in mHC: limited write-back information supply and the cubic cost of dense residual mixing.

\textbf{2) Expanded Hyper-Connections.} We propose the first HC-family method to achieve meaningful expansion beyond $N{=}4$, through two coordinated designs: temporal feature augmentation for richer write-back, and sparse residual updates for affordable residual mixing.

\textbf{3) Large-scale validation.} We validate xHC on 18B and 28B MoE models and observe strong and consistent downstream improvements across both scales. Scaling laws and $N$-sweeps further demonstrate systematic compute and expansion efficiency, while experiments with Muon confirm that xHC remains effective beyond AdamW.

\textbf{4) Efficient deployment.} We introduce xHC-Flash and fused kernel implementations to reduce memory traffic and kernel launch overhead in large-$N$ residual-stream expansion. xHC-Flash ($N=16$, $k=4$) achieves per-sublayer memory traffic comparable to mHC at $N{=}4$ ($40C$ vs.\ $34C$) while retaining most of the performance gains of full xHC.

\vspace{-3pt}
\section{Related Work}
\vspace{-3pt}

\textbf{Cross-Layer Information Flow.}
Residual connections~\citep{he2016deep} are the default mechanism for carrying token representations across layers in Transformer language models~\citep{vaswani2017attention}. Most LLM architectures~\citep{liu2025deepseek32,grattafiori2024llama,yang2025qwen3} keep this residual state as a single stream, while improving its optimization or weighting through gates, learned scaling, or depth-dependent rescaling~\citep{srivastava2015highway,layerscale,bachlechner2021rezero,deepnorm}. Other methods improve cross-layer access or feature reuse~\citep{densenet,pagliardini2024denseformer,AttnRes,MoDA}, but operate over previous activations rather than expanding the persistent residual state. Closest to our work, Hyper-Connections (HC)~\citep{zhu2024hyper} expand the residual stream into $N$ learnable streams, and Manifold-Constrained HC (mHC)~\citep{DS_mHC} stabilizes this formulation at scale via Sinkhorn-normalized residual mixing, achieving the strongest performance among HC-family methods. Other HC variants~\citep{yang2026mhclite,sHC} improve expressivity or memory efficiency, but existing HC-family methods largely operate at $N{=}4$, leaving open whether residual-stream expansion can become a true scaling axis for LLM pre-training. xHC bridges this gap by making larger-$N$ expansion both meaningful and affordable.

\textbf{Sparse Computation in Language Models.}
Sparsity is widely used to scale language models efficiently. Mixture-of-Experts (MoE) applies sparsity over computation: each token is routed to a subset of experts, decoupling total parameters from per-token FLOPs~\citep{Outrageouslytopk1,GSShardtopk2,dai2024deepseekmoe,liu2025deepseek32,team2025kimi2}. Sparse attention instead sparsifies over tokens, restricting each query to attend to a subset of the sequence and reducing long-context attention cost~\citep{child2019generating,beltagy2020longformer,NSA,liu2025deepseek32}. xHC applies sparsity on a different dimension: \emph{residual streams}. It activates only $k$ out of $N$ streams for residual mixing and write-back. This reduces the dominant residual-mixing cost from scaling with all $N$ streams to scaling with only the $k$ active streams, making large-$N$ residual-memory expansion affordable.

\vspace{-3pt}
\section{Method}
\vspace{-3pt}
\label{sec:method}

\begin{figure*}[t]
    \centering
    \includegraphics[width=\textwidth]{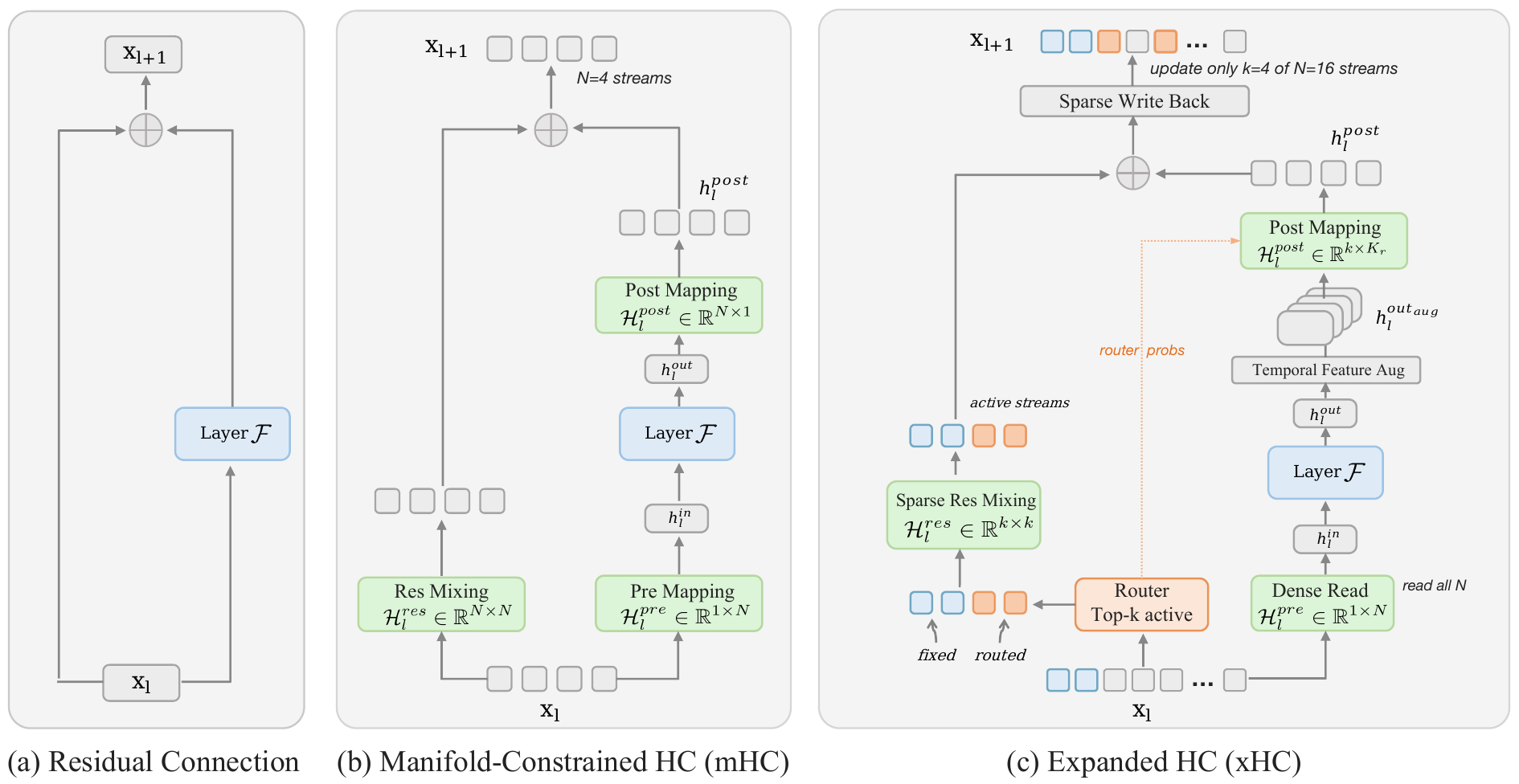}
    \caption{\textbf{Overview of xHC.}
    (a) A standard Transformer layer maintains a single residual stream.
    (b) mHC expands the residual state into $N{=}4$ streams with dense residual mixing and write-back.
    (c) xHC scales to $N{=}16$ with only $k{=}4$ active streams: it reads all streams, applies the sublayer $\mathcal{F}$ (Attn/MLP), augments MLP outputs, and sparsely writes back to selected streams. Blue/orange streams denote fixed/routed active streams. Restricting residual mapping to the $k$ active streams reduces the dominant residual-mapping generation cost from $O(N^3C)$ to $O(k^3C)$.}
    \label{fig:architecture}    
    \vspace{-10pt}
\end{figure*}

We first recap the HC formulation and its manifold-constrained variant (\S\ref{sec:prelim}), then analyze why scaling the expansion rate saturates in existing designs (\S\ref{sec:bottlenecks}). We then present xHC (\S\ref{sec:xhc_overview}; Figure~\ref{fig:architecture}), which uses two coordinated designs to make large-$N$ expansion effective and affordable: temporal feature augmentation enriches the write-back signal (\S\ref{sec:enrich}), while a sparse residual-stream architecture reduces the cost of large-$N$ residual mixing (\S\ref{sec:sparse}).

\vspace{-3pt}
\subsection{Preliminaries}
\label{sec:prelim}
\vspace{-3pt}
\textbf{Hyper-Connections.}
Standard residual connections~\citep{he2016deep} maintain a single residual stream across layers, limiting the capacity of cross-layer information flow. Hyper-Connections (HC)~\citep{zhu2024hyper} generalize this design by maintaining $N$ parallel residual streams, yielding significant pre-training improvements with modest additional FLOPs. Let $X_l = (x_{l,1}, \dots, x_{l,N})^\top \in \mathbb{R}^{N \times C}$ denote the multi-stream state at layer $l$, where $C$ is the hidden dimension. HC introduces three learnable mappings per layer: a \emph{pre-mapping} $\mathcal{H}_l^{\mathrm{pre}} \in \mathbb{R}^{1 \times N}$ that aggregates streams into a single input, a \emph{post-mapping} $\mathcal{H}_l^{\mathrm{post}} \in \mathbb{R}^{N \times 1}$ that distributes the layer output back onto streams, and a \emph{residual mapping} $\mathcal{H}_l^{\mathrm{res}} \in \mathbb{R}^{N \times N}$ that mixes streams. The single-layer update is:
\begin{equation}
\label{eq:hc}
X_{l+1} = \mathcal{H}_l^{\mathrm{res}}\, X_l \;+\; \mathcal{H}_l^{\mathrm{post}}\, \mathcal{F}\!\left(\mathcal{H}_l^{\mathrm{pre}}\, X_l,\; \mathcal{W}_l\right),
\end{equation}
where $\mathcal{F}(\cdot, \mathcal{W}_l)$ is the layer function (Attention or MLP) with parameters $\mathcal{W}_l$. In the dynamic variant, all three mappings are predicted from the current state via lightweight projections gated by $\tanh$.

\textbf{Manifold-Constrained HC.}
When $\mathcal{H}_l^{\mathrm{res}}$ is unconstrained, the composed mapping $\prod_{l} \mathcal{H}_l^{\mathrm{res}}$ across depth can amplify or attenuate signals, breaking the identity mapping property that keeps deep residual training stable. mHC~\citep{DS_mHC} addresses this by projecting $\mathcal{H}_l^{\mathrm{res}}$ onto the Birkhoff polytope of doubly stochastic matrices via the Sinkhorn--Knopp algorithm:
\begin{equation}
\label{eq:mhc_sinkhorn}
\mathcal{H}_l^{\mathrm{res}} = \mathrm{SK}\!\left(\exp\!\left(\alpha_l^{\mathrm{res}} \cdot (x_l' W_l^{\mathrm{res}}) + b_l^{\mathrm{res}}\right)\right) \in \mathbb{R}^{N \times N},
\end{equation}
where $x_l' \in \mathbb{R}^{NC}$ is the normalized flattened multi-stream state, $W_l^{\mathrm{res}} \in \mathbb{R}^{NC \times N^2}$ predicts the residual-mixing logits, $\alpha_l^{\mathrm{res}}$ and $b_l^{\mathrm{res}}$ are learnable scale and bias, and $\mathrm{SK}(\cdot)$ denotes Sinkhorn normalization~\citep{sinkhorn1967concerning} that enforces row and column sums to one. This constraint preserves stable information propagation across depth, enabling practical large-scale deployment at $N{=}4$ with significant performance gains.

\vspace{-3pt}
\subsection{Why Scaling $N$ Saturates in mHC}
\label{sec:bottlenecks}
\vspace{-3pt}

Understanding this saturation is essential for turning residual-stream expansion from a small-$N$ improvement into a practical scaling axis.
Prior results~\citep{zhu2024hyper} show that early residual-stream expansion is highly effective: increasing $N$ from 1 to 4 brings substantial gains at modest FLOPs cost (less than 2\%).
This suggests that scaling $N$ could improve model performance by increasing residual memory, providing an axis orthogonal to width and depth.
However, existing HC-family methods~\citep{zhu2024hyper,yang2026mhclite,sHC,DS_mHC} typically stop at $N{=}4$, leaving open whether this axis remains meaningful at larger $N$.
To answer this, we sweep $N \in \{2,4,8,16,32\}$ in mHC under matched training recipes.
As shown in Figure~\ref{fig:teaser_expansion}, scaling beyond $N{=}4$ quickly runs into diminishing returns: increasing $N$ from 4 to 16 reduces loss by only 0.006, while training FLOPs increase by 32\%.
This unfavorable benefit-cost tradeoff motivates us to examine what prevents larger $N$ from translating into useful capacity.
We identify two bottlenecks behind this saturation.

\textbf{Information Supply Bottleneck.}
\label{sec:info_bottleneck}
In mHC, each of the $N$ streams accumulates a distinct weighted combination of historical layer outputs. However, the write-back to stream $i$ at layer $l$ takes the form:
\begin{equation}
\label{eq:writeback}
\Delta x_{l,i} = h_{l,i}^{\mathrm{post}} \cdot \mathrm{out},
\end{equation}
where $h_{l,i}^{\mathrm{post}}$ is a scalar and $\mathrm{out} \in \mathbb{R}^{C}$ is the layer output. Although $h_{l,i}^{\mathrm{post}}$ is input-dependent and stream-specific, the newly injected information at each layer is spanned by only one write-back component, $\mathrm{out}$: different streams can assign different weights to this component, but cannot draw from different components. 
This can be sufficient at small $N$, but as $N$ grows, additional streams need more diverse write-back components to form non-redundant histories. Without such diversity, extra streams become increasingly redundant.
Section~\ref{sec:ablation} provides evidence for this bottleneck.

\textbf{Cost Bottleneck.}
\label{sec:cost_bottleneck}
The second bottleneck is on the cost side. In mHC, generating $\mathcal{H}_l^{\mathrm{res}} \in \mathbb{R}^{N \times N}$ from the $NC$-dimensional state requires predicting $N^2$ coefficients, leading to an $O(N^3 C)$ input-dependent projection cost. This makes large-$N$ expansion increasingly expensive even when the additional performance gain is marginal. Concretely, mHC at $N{=}16$ adds roughly 32\% extra FLOPs relative to its $N{=}4$ setting for only limited loss reduction on 2.5B MoE, as shown in Figure~\ref{fig:teaser_expansion}.

The two bottlenecks compound: the information bottleneck limits the \emph{benefit} of larger $N$, while the cost bottleneck increases its \emph{price}. Together they collapse the return on investment of scaling $N$.

\vspace{-3pt}
\subsection{xHC: Expanded Hyper-Connections}
\label{sec:xhc_overview}
\vspace{-3pt}

xHC addresses both bottlenecks to make expansion rate $N$ a practical scaling axis (Figure~\ref{fig:architecture}). It enriches write-back signals with local contextual features and adopts an asymmetric sparse residual-stream architecture: only $k$ active streams undergo residual mixing and write-back, while dense read keeps every layer connected to all $N$ streams. This reduces the dominant $\mathcal{H}_l^{\mathrm{res}}$ generation cost from $O(N^3 C)$ to $O(k^3 C)$, with $k{=}4$ and $N=16$ in our main setting. Other implementation details follow mHC unless stated otherwise.

\vspace{-3pt}
\subsubsection{Enriching the Write-Back Signal}
\vspace{-3pt}
\label{sec:enrich}

\textbf{Temporal Feature Augmentation.}
As analyzed in \S\ref{sec:info_bottleneck}, large-$N$ expansion saturates because each layer provides only a single write-back component to all streams. A direct fix would be to compute a separate output for each stream, but this would multiply the layer FLOPs by $N$. Instead, xHC enriches the write-back basis by borrowing low-cost local contextual information from neighboring tokens. Because autoregressive prediction already conditions on surrounding context, neighboring hidden outputs are semantically compatible with the current token and can be integrated into its write-back signal. We aggregate this local information with lightweight causal depthwise 1D convolutions, which preserve autoregressive order and add small overhead~\citep{gulati2020conformer,wu2019payconv,gu2023mamba}.

To describe local convolution, we temporarily restore the sequence dimension and write $\mathrm{out} \in \mathbb{R}^{S \times C}$, where $S$ is the sequence length. We apply $r$ causal depthwise convolutions with kernel sizes $\{\kappa_1,\dots,\kappa_r\}$ to the layer output. Different kernel sizes capture neighboring-token information at different contextual ranges, providing multi-granularity write-back components that expand the write-back basis beyond a single layer output. We concatenate these components with the original output:
\begin{equation}
\label{eq:aug}
\mathrm{out}_{\mathrm{aug}} =
\left[\,\mathrm{out};\; \mathrm{DWConv}_{\kappa_1}(\mathrm{out});\; \dots;\; \mathrm{DWConv}_{\kappa_r}(\mathrm{out})\,\right]
\in \mathbb{R}^{S \times K_r \times C},
\end{equation}
where $K_r=r+1$. In our main setting, $r=3$ with kernel sizes $\{4,8,12\}$, so $K_r=4$. The convolutions are per-channel and causal, adding only $C\sum_{j=1}^{r} \kappa_j$ parameters per layer. We apply temporal augmentation only after MLP (MoE FFN) layers: attention already mixes positions, and adding temporal augmentation after attention empirically destabilizes training.

\textbf{Gram--Schmidt Orthogonalization.}
Because depthwise convolutions operate channel-wise, their outputs can retain a strong component aligned with the original layer output. If these correlated components are directly combined by $\mathcal{H}_l^{\mathrm{post}}$, the augmented write-back signal may contain redundant components and amplify the original direction in an uncontrolled manner, making write-back scaling less predictable at large $N$. We therefore apply modified Gram--Schmidt orthogonalization~\citep{GS} over the $K_r$ components. Let $g_j=\mathrm{DWConv}_{\kappa_j}(\mathrm{out})$ denote the $j$-th convolutional component, and initialize $v_1=\mathrm{out}$. For $j=1,\dots,r$, we compute:
\begin{equation}
\label{eq:gs}
v_{j+1}
=
g_j -
\sum_{i=1}^{j}
\frac{\langle g_j, v_i \rangle}{\langle v_i, v_i \rangle}
v_i .
\end{equation}
After orthogonalization, we redefine $\mathrm{out}_{\mathrm{aug}} = [v_1;\dots;v_{K_r}] \in \mathbb{R}^{S \times K_r \times C}$ and use these orthogonalized components for all subsequent write-back operations.
The projection is computed per token along the $C$-dimensional space and adds small overhead.

\vspace{-3pt}
\subsubsection{Sparse Residual-Stream Architecture}
\vspace{-3pt}
\label{sec:sparse}

The cost bottleneck comes from residual mixing over all $N$ streams. xHC reduces this cost by updating only $k$ active streams out of $N$, while keeping dense read access to the full $N$-stream state. We describe the forward pass in three steps: routing, reading, and writing.

\textbf{Stream Routing.}
\label{sec:router}
A router selects the $k$ streams to update at each sublayer. Let $\tilde{x}_l \in \mathbb{R}^{NC}$ denote the LayerNorm-normalized~\citep{layernorm} flattened multi-stream state used by the router. The router observes the full $N$-stream state and produces per-stream importance scores:
\begin{equation}
\label{eq:router}
s = \sigma\!\left(\tilde{x}_l W_r\right) \in \mathbb{R}^{N},
\end{equation}
where $W_r \in \mathbb{R}^{NC \times N}$ is a learned projection and $\sigma$ is the sigmoid function. We use sigmoid rather than softmax to reduce winner-take-all routing. For stability, we use a fixed-plus-routed scheme: $m$ streams are always active with routing weight 1, while the remaining $k-m$ active streams are selected by TopK routing over the non-fixed streams~\citep{GSShardtopk2,Outrageouslytopk1}. Let $\mathcal I=(\mathcal I_1,\dots,\mathcal I_k)$ denote the selected active stream indices:
\begin{equation}
\label{eq:router_topk}
\mathcal{I}
=
\mathrm{FixedIdx}
\cup
\mathrm{TopK}_{i>m}(s_i, k-m),
\qquad
p_j =
\begin{cases}
1, & \mathcal I_j \in \mathrm{FixedIdx},\\
s_{\mathcal I_j}, & \mathcal I_j \notin \mathrm{FixedIdx},
\end{cases}
\quad j=1,\dots,k .
\end{equation}
Although the projection produces scores for all $N$ streams, the scores of the fixed streams are ignored in TopK selection. This full-score projection makes it easier to fuse router generation with the pre-mapping projection (Section~\ref{sec:infrastructure}) while guaranteeing $m$ persistent active paths.

\textbf{Dense Read.}
A naive sparse variant would sparsify both reading and writing, but this can disconnect cross-layer information flow: streams updated at one layer may not be selected for reading by the next, shortening the effective path through which routed information propagates. xHC therefore keeps the read path dense:
\begin{equation}
\label{eq:dense_read}
\mathrm{input}_l = \sum_{i=1}^{N} h_{l,i}^{\mathrm{pre}} \cdot x_{l,i}, \quad
\text{where } \mathcal{H}_l^{\mathrm{pre}} = f_{\mathrm{pre}}(X_l) \in \mathbb{R}^{1 \times N}.
\end{equation}
Thus every layer can access the full $N$-stream state, preserving cross-layer information flow even though only $k$ streams are updated. Ablations in Section~\ref{sec:ablation} confirm the importance of dense read.

\textbf{Sparse Residual Update.}
In mHC, $\mathcal{H}_l^{\mathrm{res}}$ and $\mathcal{H}_l^{\mathrm{post}}$ are generated from the full $N$-stream state. xHC instead applies the same generators only to the active state $X_{\mathrm{active}}$:
\begin{align}
\mathcal{H}_l^{\mathrm{res}} &= \mathrm{SK}\!\left(f_{\mathrm{res}}(X_{\mathrm{active}})\right) \in \mathbb{R}^{k \times k}, \label{eq:sparse_hres} \\
\mathcal{H}_l^{\mathrm{post}} &= f_{\mathrm{post}}(X_{\mathrm{active}}) \in \mathbb{R}^{k \times K_r}. \label{eq:sparse_hpost}
\end{align}
Here, $f_{\mathrm{res}}$ and $f_{\mathrm{post}}$ follow the residual and post-mapping generators of mHC, but operate on the active streams. The post-mapping $\mathcal{H}_l^{\mathrm{post}}$ generalizes the mHC form from $\mathbb{R}^{N \times 1}$ to $\mathbb{R}^{k \times K_r}$, so that each active stream can independently combine the $K_r$ augmented write-back components. This reduces the dominant residual-mapping generation cost from $O(N^3 C)$ to $O(k^3 C)$ while retaining the Sinkhorn constraint on the active stream subset.

The active write-back signal is computed by contracting
$\mathcal{H}_l^{\mathrm{post}}$ with the augmented components:
\begin{equation}
\label{eq:sparse_write_delta}
\Delta X_{\mathrm{active}, j}
=
p_j \sum_{r=1}^{K_r}
\mathcal{H}_{l,j,r}^{\mathrm{post}}
\, \mathrm{out}_{\mathrm{aug},r},
\quad j=1,\dots,k,
\end{equation}
where the token dimension is omitted for clarity. The active streams are then updated by:
\begin{equation}
\label{eq:sparse_write}
X_{\mathrm{active}}^{\mathrm{new}}
=
\mathcal{H}_l^{\mathrm{res}} X_{\mathrm{active}}
+
\Delta X_{\mathrm{active}} .
\end{equation}
Here $p_j$ is the routing weight for the $j$-th active stream: fixed streams use $p_j=1$, while routed streams use their sigmoid scores. We apply $p_j$ only to the new write-back term, leaving residual mixing ungated. Updated streams are scattered back to the full state, while non-active streams carry forward unchanged for future dense reads.

\vspace{-3pt}
\subsubsection{Why Both Designs Are Needed}
\vspace{-3pt}
\label{sec:synergy}

The two designs address complementary bottlenecks. Temporal feature augmentation makes additional streams more informative, but alone leaves dense residual mixing expensive. Sparse residual updates make large $N$ affordable, but alone leave the write-back signal information-limited. Together, they make large-$N$ expansion both meaningful and affordable, with ablations in Section~\ref{sec:ablation} confirming that neither design alone recovers the full benefit of xHC.

\begin{algorithm}[t]
\caption{Forward pass of one xHC sublayer}\label{alg:xhc}
\begin{algorithmic}[1]
\REQUIRE $X \in \mathbb{R}^{N \times C}$ (multi-stream state), sublayer $\mathcal{F}$ (Attention/MLP)
\STATE
\STATE \textbf{/* Stream routing */}
\STATE $s \leftarrow \sigma(\mathrm{LN}(\mathrm{flatten}(X))\, W^{\mathrm{r}})$, \quad $\mathcal{H}^{\mathrm{pre}} \leftarrow f_{\mathrm{pre}}(X)$ \hfill $// \;\mathbb{R}^N,\; \mathbb{R}^N$
\STATE $\mathcal{I}, p \leftarrow \mathrm{Route}(s; m,k)$ \hfill $// \;m$ fixed and $k{-}m$ routed streams
\STATE $X_{\mathrm{act}} \leftarrow \mathrm{Gather}(X, \mathcal{I})$ \hfill $// \;\mathbb{R}^{k \times C}$
\STATE
\STATE \textbf{/* Mapping generation (from active streams) */}
\STATE $\mathcal{H}^{\mathrm{res}},\, \mathcal{H}^{\mathrm{post}} \leftarrow f_{\mathrm{res}}(X_{\mathrm{act}}),\; f_{\mathrm{post}}(X_{\mathrm{act}})$ \hfill $// \;\mathbb{R}^{k \times k},\; \mathbb{R}^{k \times K_r}$
\STATE
\STATE \textbf{/* Dense read $\to$ sublayer $\to$ temporal feature augmentation */}
\STATE $\mathrm{out} \leftarrow \mathcal{F}\!\left(\textstyle\sum_{i=1}^{N} \mathcal{H}^{\mathrm{pre}}_i \cdot x_i\right)$ \hfill $// \;\mathbb{R}^{C}$
\IF{$\mathcal{F}$ is MLP}
    \STATE $\mathrm{out}_{\mathrm{aug}} \leftarrow \mathrm{TempAug}(\mathrm{out})$ \hfill $// \;\mathbb{R}^{K_r \times C}$
\ELSE
    \STATE $\mathrm{out}_{\mathrm{aug}} \leftarrow \mathrm{out}$ \hfill $// \;K_r=1$
\ENDIF
\STATE
\STATE \textbf{/* Sparse write-back */}
\STATE $X_{\mathrm{act}}^{\mathrm{new}} \leftarrow \mathcal{H}^{\mathrm{res}} X_{\mathrm{act}} + p \odot (\mathcal{H}^{\mathrm{post}}\, \mathrm{out}_{\mathrm{aug}})$ \hfill $// \;\mathbb{R}^{k \times C}$
\STATE $X \leftarrow \mathrm{Scatter}(X,\; \mathcal{I},\; X_{\mathrm{act}}^{\mathrm{new}})$ \hfill $//$ inactive streams unchanged
\RETURN $X$
\end{algorithmic}
\end{algorithm}
\vspace{-3pt}
\subsubsection{Parameterization Details}
\vspace{-3pt}
\label{sec:parameterization}

We now detail the concrete parameterization of the generators $f_{\mathrm{pre}}$, $f_{\mathrm{res}}$, and $f_{\mathrm{post}}$ left implicit in Eqs.~\ref{eq:dense_read}--\ref{eq:sparse_hpost}. These follow a shared gated-projection pattern from mHC: a learnable gating scalar $\alpha$ (initialized to 0.01) scales the input-dependent term relative to a learnable bias, so that each mapping starts near its static initialization and gradually becomes dynamic. The stream router uses a direct linear projection followed by sigmoid activation (Eq.~\ref{eq:router}).

The pre-mapping is generated from the RMSNorm-normalized full state:
\begin{align}
\mathcal{H}_l^{\mathrm{pre}} &= \sigma\!\left(\alpha^{\mathrm{pre}} \cdot \mathrm{RMSNorm}(\mathrm{flatten}(X_l)) \, W^{\mathrm{pre}} + b^{\mathrm{pre}}\right), \quad W^{\mathrm{pre}} \in \mathbb{R}^{NC \times N}. \label{eq:param_pre}
\end{align}
The residual and post-mapping generators use the RMSNorm-normalized active state $\tilde{x}_{\mathrm{active}} = \mathrm{RMSNorm}(\mathrm{flatten}(X_{\mathrm{active}})) \in \mathbb{R}^{kC}$:
\begin{align}
\mathcal{H}_l^{\mathrm{res}} &= \mathrm{SK}\!\left(\exp\!\left(\alpha^{\mathrm{res}} \cdot \mathrm{mat}_{k \times k}(\tilde{x}_{\mathrm{active}} \, W^{\mathrm{res}}) + b^{\mathrm{res}}\right)\right), \quad W^{\mathrm{res}} \in \mathbb{R}^{kC \times k^2}, \label{eq:param_res} \\
\mathcal{H}_l^{\mathrm{post}} &= 2\sigma\!\left(\alpha^{\mathrm{post}} \cdot \mathrm{mat}_{k \times K_r}(\tilde{x}_{\mathrm{active}} \, W^{\mathrm{post}}) + b^{\mathrm{post}}\right), \quad W^{\mathrm{post}} \in \mathbb{R}^{kC \times kK_r}. \label{eq:param_post}
\end{align}
The $2\sigma(\cdot)$ scaling in the post-mapping allows coefficients in $[0, 2]$, enabling both attenuation and mild amplification of individual write-back components. The sigmoid in the pre-mapping constrains aggregation weights to $[0, 1]$. For the attention sublayer (no temporal augmentation), $K_r = 1$ and the post-mapping reduces to $\mathbb{R}^{k \times 1}$. The temporal convolution branches each have a kernel $w_j \in \mathbb{R}^{\kappa_j \times C}$, computing $g_j[t] = \sum_{i=0}^{\kappa_j - 1} w_j[i] \odot \mathrm{out}[t{-}i]$ channel-wise; with $r{=}3$ branches of kernel sizes $\{4,8,12\}$, this adds only $24C$ parameters per MLP sublayer. Algorithm~\ref{alg:xhc} summarizes the forward pass of a single xHC sublayer.

\vspace{-2pt}
\section{Experiments}
\vspace{-5pt}
\label{sec:experiments}

We evaluate xHC through language model pre-training experiments across model scales, compute budgets, expansion rates, ablation settings, and optimizer choices. We first compare xHC against mHC and the vanilla residual baseline on 18B and 28B MoE models, then study compute efficiency through scaling-law experiments. Next, $N$-sweep and ablation studies examine whether larger expansion rates remain beneficial under xHC and whether the two designs address the large-$N$ bottlenecks. Finally, we test compatibility with the Muon optimizer.
\begin{table*}[t]
    \centering
    \small
    \caption{Downstream evaluation on 18B and 28B MoE models. All numbers are scores (\%; higher is better). mHC uses $N{=}4$, and xHC uses $N{=}16$, $k{=}4$; all methods use comparable training FLOPs.}
    \label{tab:main_results}
    \begin{tabular}{l ccc ccc}
        \toprule
        & \multicolumn{3}{c}{\textbf{18B}} & \multicolumn{3}{c}{\textbf{28B}} \\
        \cmidrule(lr){2-4} \cmidrule(lr){5-7}
        Benchmark & Vanilla & mHC & xHC & Vanilla & mHC & xHC \\
        \midrule
        \multicolumn{7}{l}{\textit{Language Understanding \& Knowledge}} \\
        \quad MMLU        & 48.9 & 54.7 & \textbf{57.2} & 54.6 & 56.8 & \textbf{60.5} \\
        \quad MMLU-Pro    & 21.1 & 27.4 & \textbf{29.7} & 30.1 & 34.9 & \textbf{36.0} \\
        \quad MMLU-Redux  & 46.4 & 49.9 & \textbf{52.8} & 50.6 & 53.9 & \textbf{56.4} \\
        \midrule
        \multicolumn{7}{l}{\textit{Reasoning \& Mathematics}} \\
        \quad BBH           & 32.4 & 33.7 & \textbf{39.5} & 41.7 & \textbf{43.6} & 43.4 \\
        \quad CommonsenseQA  & 54.6 & 56.6 & \textbf{60.9} & 60.5 & 63.9 & \textbf{69.6} \\
        \quad ARC-Challenge  & 55.7 & 66.3 & \textbf{72.2} & 70.8 & 74.9 & \textbf{77.7} \\
        \quad GSM8K & 37.7 & 44.5 & \textbf{48.4} & 50.3 & 56.3 & \textbf{59.2} \\
        \midrule
        \multicolumn{7}{l}{\textit{Code}} \\
        \quad HumanEval   & 25.6 & 23.2 & \textbf{29.3} & 27.4 & 26.8 & \textbf{31.1} \\
        \quad LCBench     &  9.9 & 12.2 & \textbf{14.6} & 15.1 & 14.8 & \textbf{17.9} \\
        \midrule
        \multicolumn{7}{l}{\textit{Chinese}} \\
        \quad CMMLU       & 42.7 & 47.6 & \textbf{50.4} & 47.6 & 50.1 & \textbf{53.4} \\
        \quad CEval       & 44.5 & 48.8 & \textbf{52.4} & 50.2 & 51.2 & \textbf{54.9} \\
        \quad C3          & 67.1 & 72.7 & \textbf{78.3} & 75.2 & 78.7 & \textbf{82.5} \\
        \midrule
        \textbf{Average}  & 40.6 & 44.8 & \textbf{48.8} & 47.8 & 50.5 & \textbf{53.6} \\
        \bottomrule
    \end{tabular}
    \vspace{-9pt}
\end{table*}

\vspace{-6pt}
\subsection{Experimental Setup}
\vspace{-5pt}

\textbf{Training setup.}
We conduct experiments in the language model pre-training setting using Mixture-of-Experts (MoE) Transformer models. Unless otherwise stated, we apply xHC and mHC to the same DeepSeekMoE-style~\citep{dai2024deepseekmoe} Transformer backbone. Specifically, the backbone uses grouped-query attention (GQA)~\citep{ainslie2023gqa} and 144 experts with top-8 routing, where each expert is implemented as a SwiGLU-based FFN~\citep{shazeer2020glu}. We report main downstream results at two MoE scales: an 18B-total, 1.7B-activated model and a 28B-total, 2.7B-activated model. We additionally use a 10B MoE model for ablations and a 2.5B MoE model for $N$-sweep experiments. Unless otherwise specified, xHC uses $N{=}16$ total streams with $k{=}4$ active streams. All models are trained with a context length of 8192 tokens on a data mixture covering English, Chinese, code, mathematics, and reasoning corpora. Unless otherwise specified, all experiments use AdamW~\cite{adamw} as the training optimizer. Full hyperparameters and training details are provided in Appendix~\ref{app:training_details}.

\textbf{Baselines.}
We compare xHC primarily against mHC ($N{=}4$)~\citep{DS_mHC}, which has comparable training FLOPs to xHC and serves as the strongest HC-family method at scale. We also include a standard residual-stream Transformer~\citep{vaswani2017attention,he2016deep}
as the vanilla baseline.

\textbf{Downstream Evaluation.}
We evaluate on a benchmark suite covering four aspects of LLM performance: language understanding and knowledge (MMLU~\citep{MMLU}, MMLU-Pro~\citep{wang2024mmlupro}, MMLU-Redux~\citep{redux}); reasoning and mathematics (BBH~\citep{BBH}, CommonsenseQA~\citep{talmor2019commonsenseqa}, ARC-Challenge~\citep{ARCC}, GSM8K~\citep{gsm8k}); code generation and understanding (HumanEval~\citep{humaneval}, LCBench~\citep{2023opencompass}); and Chinese language understanding (CMMLU~\citep{li2024cmmlu}, CEval~\citep{ceval}, C3~\citep{c3}). Together, these benchmarks provide a compact yet diverse view of whether improved residual-stream scaling translates into stronger generalization after pre-training. Evaluation protocols and shot settings are provided in Appendix~\ref{app:training_details}.

\vspace{-3pt}
\subsection{Main Results}
\vspace{-3pt}

We compare xHC against mHC and the vanilla residual baseline on 18B and 28B MoE models. Table~\ref{tab:main_results} summarizes downstream results under comparable training-FLOPs budgets, with xHC adding only 3.0\% training FLOPs over the vanilla baseline at 28B.

xHC outperforms both mHC and the vanilla baseline across both scales. At 18B, the average score improves from 44.8 with mHC to 48.8 with xHC, with representative gains on ARC-Challenge (+5.9), BBH (+5.8), C3 (+5.6), and HumanEval (+6.1). At 28B, the average score improves from 50.5 to 53.6 (+3.1) over mHC, with representative gains on CommonsenseQA (+5.7), HumanEval (+4.3), C3 (+3.8), and MMLU (+3.7). The consistent improvements on the 28B MoE model further demonstrate that xHC remains effective in larger pre-training regimes.

\begin{figure*}[t]
    \centering
    \includegraphics[width=0.78\textwidth]{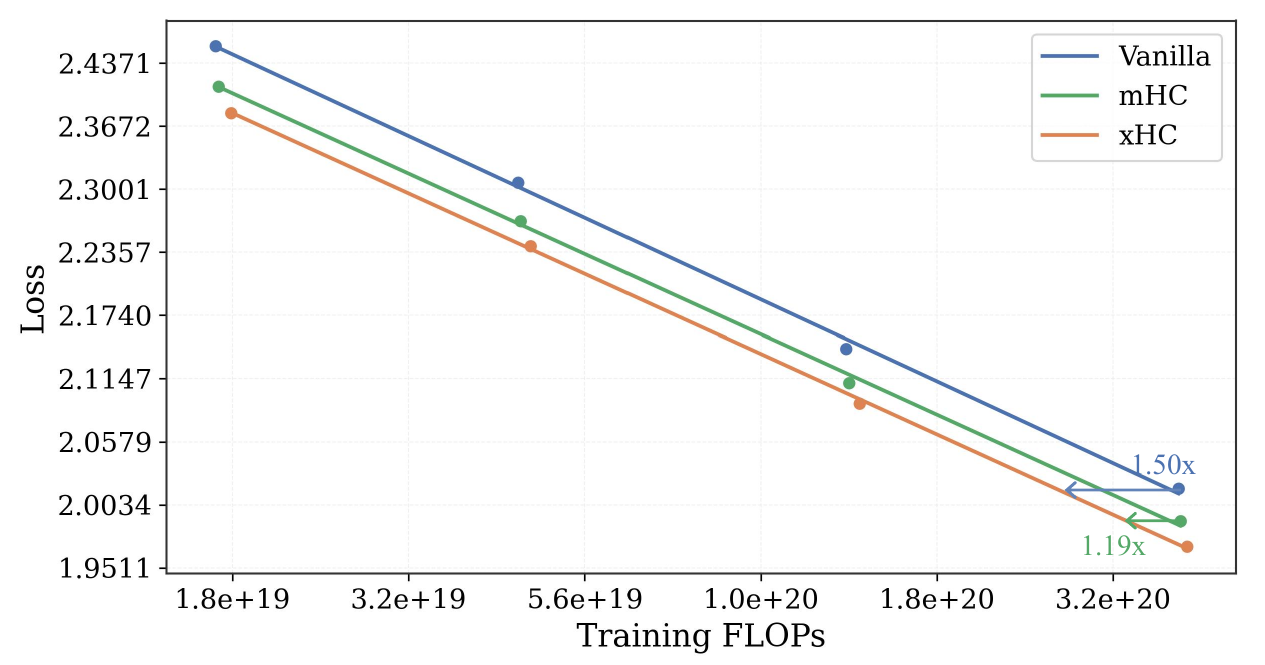}
    \caption{Scaling-law comparison. xHC traces a consistently lower loss curve than both mHC and the vanilla residual baseline across training compute.}
    \label{fig:scaling_law}
    \vspace{-15pt}
\end{figure*}

\vspace{-3pt}
\subsection{Scaling Laws}
\vspace{-3pt}
\label{sec:scaling_law}

The downstream results above establish the advantage of xHC at two discrete scales. We next ask whether this advantage persists across a broader compute range. To fit scaling laws~\citep{kaplan2020scaling}, we train a separate suite of four models for each method under matched recipes, spanning training compute from $1.7 \times 10^{19}$ to $4.0 \times 10^{20}$ FLOPs. We fit a shifted power law~\citep{hoffmann2022chinchilla,henighan2020scaling}:
\begin{equation}
\label{eq:scaling_law}
\mathcal{L}(C) = A C^{-\alpha} + E,
\end{equation}
where $C$ denotes training FLOPs, $A$ and $\alpha$ are fitted parameters, and $E=0.72$ is the estimated irreducible loss. We fit by linear regression in log space and plot the resulting curves with the offset $E$ included. Detailed settings are provided in Appendix~\ref{app:scaling_law}.

Figure~\ref{fig:scaling_law} shows that xHC traces a lower fitted loss curve than both mHC and the vanilla baseline across the measured compute range. At the largest compute point, xHC reduces loss by about 1.1\% relative to mHC and 2.4\% relative to the vanilla baseline. To quantify this advantage, we use the losses achieved by the largest vanilla baseline and mHC models as representative targets, and read from the fitted xHC curve the compute required to match each target. The resulting matched-loss comparison shows that the baseline and mHC require about $1.50\times$ and $1.19\times$ the compute of xHC, respectively. This comparison, together with the consistently lower fitted curve, indicates that xHC provides a systematic compute-efficiency improvement across the measured model family, rather than only improving specific downstream evaluation points.

\vspace{-3pt}
\subsection{Scaling the Expansion Rate}
\vspace{-3pt}
\label{sec:n_scaling}

We evaluate whether xHC turns the expansion rate $N$ into a meaningful scaling axis. We sweep $N \in \{2,4,8,16\}$ in xHC on a 2.5B MoE model (details in Appendix~\ref{app:training_details}), and compare against mHC at matched expansion rates. In mHC, all $N$ streams participate in dense residual mixing, so the residual-mixing overhead grows rapidly with $N$. In xHC, increasing $N$ enlarges residual-memory capacity while the sparse update cost remains controlled by $k$. Figure~\ref{fig:teaser_expansion} plots language modeling loss during training against training FLOPs as $N$ increases. Under xHC, loss decreases consistently from $N{=}2$ to $N{=}16$, with each doubling of $N$ yielding a clear improvement at small additional FLOPs. In contrast, mHC shows rapid saturation: increasing $N$ from 4 to 16 reduces loss by only 0.006 while increasing training FLOPs by 32\%. Over the same range, xHC reduces loss by 0.012 with only 4\% additional FLOPs. These results show that xHC substantially improves the benefit-cost tradeoff of residual-stream expansion, making larger $N$ useful under a practical compute budget.

\vspace{-3pt}
\subsection{Ablation Study}
\vspace{-3pt}
\label{sec:ablation}

Ablation studies use a 10B MoE model trained under a matched data budget, providing a controlled setting for isolating the effect of each xHC design choice. Model configurations and training details are provided in Appendix~\ref{app:training_details}. We report validation loss on the Pile test set~\citep{gao2020pile}.

\begin{figure*}[t]
    \centering
    \begin{minipage}[t]{0.69\textwidth}
    \vspace{0pt}
        \centering
        \captionof{table}{Ablation study. Parentheses show extra training FLOPs over vanilla baseline. ``Sparse'' is the sparse architecture, ``D. Read'' is Dense Read, and ``Fixed'' is the number of always-active streams.}
        \label{tab:ablation}
        \resizebox{\linewidth}{!}{
        \begin{tabular}{l ccc | cccc | c}
            \toprule[1pt]
            \textbf{Method} & $N$ & Temp Aug. & Sparse & D. Read & $k$ & Fixed & Router & Val. Loss $\downarrow$ \\
            \midrule
            (1) Vanilla baseline                    & -  & -      & -      & -      & -   & -   & -       & 2.029 \\
            (2) mHC (+0.6\%)                    & 4  & -      & -      & -      & -   & -   & -       & 2.004 \\
            (3) mHC (+18.8\%)                   & 16 & -      & -      & -      & -   & -   & -       & 1.998 \\
            (4) mHC w/ Temp Aug (+20.1\%)       & 16 & \cmark & -      & -      & -   & -   & -       & 1.984 \\
            (5) \textbf{xHC} (+3.3\%)           & 16 & \cmark & \cmark & \cmark & 4   & 2   & Sigmoid & 1.983 \\
            \midrule
            (6) w/o D.Read \& Fixed Streams     & 16 & \cmark & \cmark & \xmark & 4   & 0   & Sigmoid & 1.997 \\
            (7) w/o Dense Read                  & 16 & \cmark & \cmark & \xmark & 4   & 2   & Sigmoid & 1.985 \\
            (8) w/o Fixed Streams               & 16 & \cmark & \cmark & \cmark & 4   & 0   & Sigmoid & 1.986 \\
            (9) $k{=}2$                         & 16 & \cmark & \cmark & \cmark & 2   & 1   & Sigmoid & 1.991 \\
            (10) $k{=}8$                        & 16 & \cmark & \cmark & \cmark & 8   & 2   & Sigmoid & 1.982 \\
            \midrule
            (11) Softmax Router                 & 16 & \cmark & \cmark & \cmark & 4   & 2   & Softmax & 1.988 \\
            \bottomrule[1pt]
        \end{tabular}
        }
    \end{minipage}
    \hfill
    \begin{minipage}[t]{0.29\textwidth}
    \vspace{+3pt}
        \centering
        \includegraphics[width=\linewidth]{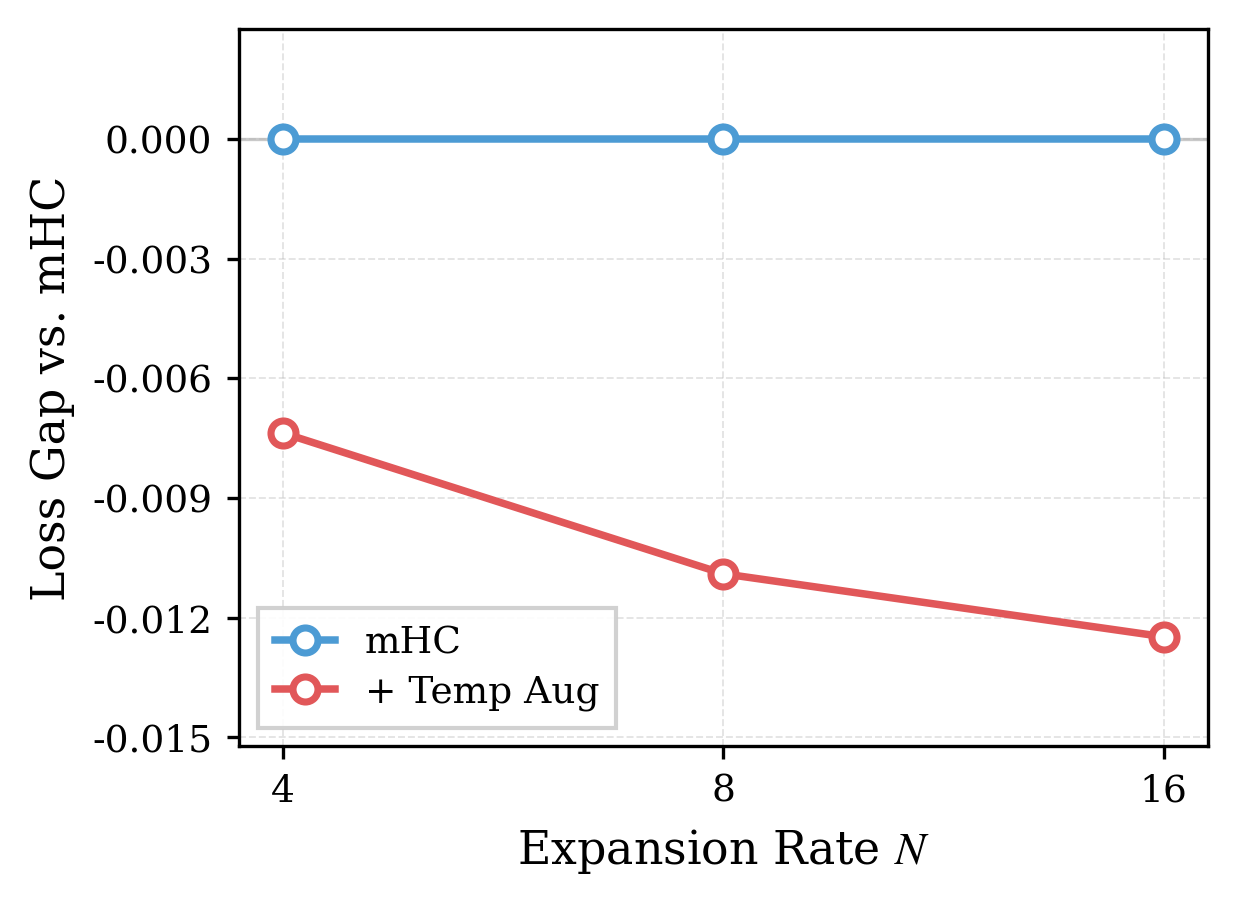}
        \captionof{figure}{Information bottleneck ablation. The gain grows with $N$.
        }
        \label{fig:info_bottleneck_ablation}
    \end{minipage}
    \vspace{-10pt}
\end{figure*}

\textbf{Incremental Construction.}
Following the motivation of xHC, we start from mHC at $N{=}16$ and incrementally add the two ingredients of xHC: temporal feature augmentation and a sparse residual-stream architecture.
Temporal feature augmentation targets the information bottleneck by enriching write-back information, while sparse residual updates target the cost bottleneck by making large-$N$ affordable.
As shown in Table~\ref{tab:ablation}~(2)--(5), adding temporal feature augmentation to mHC improves validation loss from 1.998 to 1.984, confirming the benefit of enriching the write-back signal at large $N$.
Adding the sparse residual-stream architecture then preserves this validation loss (1.983) while reducing FLOPs overhead from 20.1\% to 3.3\%, showing that sparse updates make the enriched large-$N$ residual state affordable.
Together, these results support the design of xHC: effective large-$N$ expansion requires both richer write-back information and controlled residual-update cost.

\textbf{Information Bottleneck Ablations.}
We add temporal feature augmentation alone to dense mHC at $N \in \{4,8,16\}$, using three multi-scale causal convolution branches by default.
Figure~\ref{fig:info_bottleneck_ablation} shows that the loss gap relative to mHC becomes increasingly negative as $N$ grows.
This trend is consistent with our diagnosis that the information bottleneck becomes more severe at larger expansion rates, where additional streams require richer write-back information to form non-redundant histories.

\textbf{Sparse Architecture Ablations.}
Table~\ref{tab:ablation} validates the designs that keep sparse residual updates effective.
The central risk of sparse write is information disconnection: a stream updated at one layer may not be selected at the next, breaking cross-layer propagation.
Removing both Dense Read and fixed streams exposes this risk, degrading loss to 1.997.
Dense Read mitigates this by letting every layer access all $N$ streams, so information in any stream remains visible regardless of routing; with fixed streams present, removing Dense Read increases loss from 1.983 to 1.985.
Fixed streams further stabilize sparse updates by providing guaranteed write targets; removing them increases loss from 1.983 to 1.986.
Separately, the active-stream budget $k$ controls the sparsity--quality tradeoff: $k{=}2$ under-provisions active streams (1.991), $k{=}8$ yields a marginal gain (1.982) at higher cost, and $k{=}4$ balances the two.
Finally, Sigmoid routing outperforms Softmax (1.983 vs.\ 1.988), consistent with avoiding winner-take-all routing that can leave streams persistently inactive.

\subsection{Compatibility with Muon Optimizer}
\label{sec:muon}

The experiments above use AdamW~\citep{adamw}. We further evaluate xHC under Muon~\citep{jordan2024muonjordan,liu2025muonLiu}, which uses Newton--Schulz iteration to construct orthogonalized matrix updates and has shown strong results in Transformer pre-training~\citep{team2025kimi2}. This tests whether xHC remains effective beyond AdamW.

When training with Muon, we make two practical adaptations. \textbf{First, we remove Gram--Schmidt (GS) orthogonalization from temporal feature augmentation.} Under AdamW, GS acts as a stability mechanism: the causal convolution branches are often highly correlated with the original layer output, so without orthogonalization the post-mapping can combine near-parallel components and amplify the write-back signal along the same direction, leading to activation growth and gradient-scale spikes. Muon reduces this sensitivity by applying Newton--Schulz orthogonalization to backbone matrix updates, which keeps update spectra more controlled. Moreover, GS removes parallel components in the forward pass and correspondingly projects out gradient components along these directions; under Muon, we find this extra projection unnecessary and slightly restrictive. We therefore remove GS when training xHC with Muon. Second, we apply Muon only to the backbone attention and MLP projection matrices, while \textbf{keeping xHC-specific routing and mapping projections under AdamW.} These xHC projections are highly unbalanced matrices, mapping $NC$ or $kC$ hidden features to low-dimensional stream-control outputs such as $N$ routing/pre-mapping weights, $k^2$ residual-mixing logits, and $kK_r$ post-mapping weights. Under the default $N{=}16$, $k{=}4$ setting, their output dimensions are much smaller than their input dimensions, making them less suitable for Muon's matrix orthogonalization. All other xHC components remain unchanged.

Table~\ref{tab:muon_results} reports results on the 18B MoE model under the same training recipe as Table~\ref{tab:main_results}, with Muon applied to backbone parameters and AdamW retained for xHC parameters; optimizer settings are provided in Appendix~\ref{app:training_details}. xHC maintains substantial gains over the Muon baseline across benchmarks, indicating that the architecture is compatible with Muon and remains effective beyond AdamW.

\begin{table*}[t]
    \centering
    \small
    \caption{Downstream evaluation on the 18B MoE model with Muon optimizer. ``AdamW'' and ``Muon'' columns are vanilla baselines under each optimizer; ``Muon + xHC'' applies xHC ($N{=}16$, $k{=}4$, without GS) on top of the Muon baseline.}
    \label{tab:muon_results}
    \begin{tabular}{l ccc}
        \toprule
        Benchmark & {AdamW (Tab.~\ref{tab:main_results})} & Muon & Muon + xHC \\
        \midrule
        \multicolumn{4}{l}{\textit{Language Understanding \& Knowledge}} \\
        \quad MMLU        & {48.9} & 51.3 & \textbf{56.6} \\
        \quad MMLU-Pro    & {21.1} & 27.9 & \textbf{32.0} \\
        \quad MMLU-Redux  & {46.4} & 49.0 & \textbf{52.2} \\
        \midrule
        \multicolumn{4}{l}{\textit{Reasoning \& Mathematics}} \\
        \quad BBH           & {32.4} & 36.1 & \textbf{42.2} \\
        \quad CommonsenseQA  & {54.6} & 49.1 & \textbf{61.3} \\
        \quad ARC-Challenge  & {55.7} & 63.6 & \textbf{75.3} \\
        \quad GSM8K          & {37.7} & 41.8 & \textbf{52.8} \\
        \midrule
        \multicolumn{4}{l}{\textit{Code}} \\
        \quad HumanEval   & {25.6} & 21.3 & \textbf{28.7} \\
        \quad LCBench     & {9.9}  & 11.0 & \textbf{16.0} \\
        \midrule
        \multicolumn{4}{l}{\textit{Chinese}} \\
        \quad CMMLU       & {42.7} & 45.2 & \textbf{50.2} \\
        \quad CEval       & {44.5} & 47.2 & \textbf{54.3} \\
        \quad C3          & {67.1} & 73.7 & \textbf{77.1} \\
        \midrule
        \textbf{Average}  & {40.6} & 43.1 & \textbf{49.9} \\
        \bottomrule
    \end{tabular}
\end{table*}

\section{Practical Deployment}
\label{sec:deployment}

This section examines the practical training efficiency of xHC. Although xHC introduces modest additional FLOPs, maintaining multiple residual streams increases memory traffic and can affect wall-clock throughput. We first analyze its computation and memory-access overhead, then introduce xHC-Flash to reduce repeated full-state accesses. \textbf{Notably, xHC-Flash-4sub achieves memory traffic comparable to that of mHC at $N{=}4$ while maintaining a clear performance advantage.} Finally, we describe our infrastructure design, specifically the kernel-fusion strategy used to reduce the runtime overhead of the remaining operations.

\subsection{Efficiency Analysis}
\label{sec:efficiency_analysis}

The FLOPs overhead of xHC is modest: with our default $N{=}16$, $k{=}4$ setting, xHC adds 3.0\% training FLOPs at the 28B scale. Detailed derivations of the training FLOPs and parameter overhead are provided in Appendix~\ref{app:flops_analysis}. As identified in prior work~\citep{DS_mHC}, the primary source of runtime overhead for HC-family methods~\citep{zhu2024hyper,DS_mHC} is often memory traffic rather than arithmetic FLOPs. Maintaining an $N$-stream residual state introduces additional I/O for normalization, projection, and residual mixing at every sublayer, which can directly affect training throughput.

Let $C$ denote the model hidden dimension. Table~\ref{tab:overhead_analysis} decomposes the per-token memory traffic of the residual-stream operations. Attention and MLP have asymmetric access costs because they use $K_r{=}1$ and $K_r{=}4$, respectively, and temporal augmentation is applied only to MLP. We therefore report amortized per-sublayer costs: the standard average divides the total traffic of one Attention/MLP block by two. With $N{=}16$ and $k{=}4$, xHC averages $55C$ reads and $18.5C$ writes per sublayer, for a total of $73.5C$. This is approximately $2.2\times$ the $34C$ required by mHC at its standard $N{=}4$ setting, primarily because xHC performs two full-state reads over the $NC{=}16C$ representation at each sublayer: one for mapping generation and one for dense-read aggregation. As an additional reference, mHC at the matched expansion rate of $N{=}16$ would require $130C$ per sublayer. Although sparse residual updates substantially reduce the cost of large-$N$ expansion, xHC still incurs higher memory traffic than the mHC $N=4$ configuration, motivating the efficiency optimizations introduced next.

\begin{table*}[t]
    \centering
    \small
    \caption{Per-token memory access of residual-stream operations (sublayer function I/O excluded). Totals averaged per sublayer. $^*$Once per block. $^\dagger$Includes MLP-side conv ($+C$ R, $+3C$ W) and Flash input correction ($+2C$ R, $+C$ W). $^\ddagger$Amortized over 4 sublayers.}
    \label{tab:overhead_analysis}
    \setlength{\tabcolsep}{4pt}
    \begin{tabular}{l | cc cc cc}
        \toprule
        & \multicolumn{2}{c}{\textbf{mHC}} & \multicolumn{2}{c}{\textbf{xHC} ($N{=}16,k{=}4$)} & \multicolumn{2}{c}{\textbf{xHC-Flash}} \\
        \cmidrule(lr){2-3} \cmidrule(lr){4-5} \cmidrule(lr){6-7}
        Operation & R & W & R & W & R & W \\
        \midrule
        Mapping generation & $NC$ & $N^2{+}2N$ & $NC$ & $2N$ & $NC^*$ & $3N^*$ \\
        $\mathcal{H}^{\mathrm{pre}}$ dense read & $NC{+}N$ & $C$ & $NC{+}N$ & $C$ & $(NC{+}2N)^*$ & $2C^*$ \\
        Gather active streams & -- & -- & $kC$ & $kC$ & $kC^*$ & $kC^*$ \\
        Active mapping generation & -- & -- & $kC$ & $k^2{+}kK_r$ & $kC$ & $k^2{+}kK_r$ \\
        $\mathcal{H}^{\mathrm{res}}$ residual mixing & $NC{+}N^2$ & $NC$ & $kC{+}k^2$ & $kC$ & $(kC{+}k^2)^*$ & $kC^*$ \\
        $\mathcal{H}^{\mathrm{post}}$ write-back & $C{+}N$ & $NC$ & $K_rC{+}kK_r$ & $kC$ & $K_rC{+}kK_r$ & $kC$ \\
        Residual merge and scatter & $2NC$ & $NC$ & $2kC$ & $kC$ & $2kC$ & $kC$ \\
        \midrule
        \textbf{Per-sublayer, avg.}$^\dagger$ & $21C$ & $13C$ & $55C$ & $18.5C$ & $36C$ & $15C$ \\
        \textbf{xHC-Flash-4sub, avg.}$^\ddagger$ & -- & -- & -- & -- & $26.5C$ & $13.5C$ \\
        \midrule
        \multirow{2}{*}{\textbf{Total I/O}}
        & \multicolumn{2}{c}{$130C$ ($N{=}16$)}
        & \multicolumn{2}{c}{\multirow{2}{*}{$73.5C$}}
        & \multicolumn{2}{c}{$51C$ (xHC-Flash)} \\
        & \multicolumn{2}{c}{$34C$ ($N{=}4$)}
        & \multicolumn{2}{c}{}
        & \multicolumn{2}{c}{$40C$ (xHC-Flash-4sub)} \\
        \bottomrule
    \end{tabular}
\end{table*}

This gap motivates the two optimizations developed below: xHC-Flash reduces repeated full-state accesses at the architectural level, while our fused implementation reduces the overhead of the remaining residual-stream operations.

\subsection{xHC-Flash: A Lightweight Variant}
\label{sec:xhc_flash}

To reduce xHC's memory traffic, we introduce \textbf{xHC-Flash}, which amortizes full-state operations across consecutive sublayers while preserving xHC's performance advantage over mHC. Unless otherwise stated, xHC-Flash follows the full-xHC configuration: $N{=}16$, $k{=}4$, and MLP-side $K_r{=}4$ temporal augmentation. Sharing within one block reduces the per-sublayer traffic from $73.5C$ to $51C$. Its natural four-sublayer extension, \textbf{xHC-Flash-4sub}, further reduces the traffic to $40C$. At this level, its traffic is comparable to that of mHC at $N{=}4$ ($34C$), and it retains most of xHC's performance gains (Table~\ref{tab:overhead_analysis}). Because only $k{=}4$ of $N{=}16$ streams change at each sublayer, xHC-Flash reuses routing, jointly forms sublayer-specific pre-mappings and base readouts, and updates subsequent inputs through lightweight corrections instead of repeated full-state reads.

\textbf{Key simplifications.}
xHC-Flash makes the following changes relative to the full xHC:
\begin{itemize}[leftmargin=*, itemsep=2pt, topsep=2pt]
    \item \textbf{Shared routing}: a single routing decision computed from the block-entry state is shared by the Attention and MLP sublayers. This removes the MLP-side router projection and full-state normalization, at the cost of conditioning MLP routing on the pre-Attention state.
    \item \textbf{Joint pre-mapping with separate weights}: the Attention and MLP pre-mappings, $\mathcal{H}^{\mathrm{pre,Attn}}$ and $\mathcal{H}^{\mathrm{pre,MLP}}$, are generated jointly from the block-entry state using separate projection weights, producing separate base readouts:
    \begin{equation}
    \label{eq:flash_dual_read}
    \mathrm{inp}_{\mathrm{A}}
    =
    \sum_{i=1}^{N}
    \mathcal{H}^{\mathrm{pre,Attn}}_i x_i,
    \qquad
    \mathrm{inp}_{\mathrm{M}}
    =
    \sum_{i=1}^{N}
    \mathcal{H}^{\mathrm{pre,MLP}}_i x_i.
    \end{equation}
    The separate weights preserve sublayer-specific stream aggregation, which we find important for retaining model performance. Both pre-mappings are conditioned on the block-entry state. After the Attention write-back, $\mathrm{inp}_{\mathrm{M}}$ is corrected using only the changes to the active streams (Eq.~\ref{eq:read_reuse}), avoiding a second full-state read. We refer to this exact correction as \emph{dense-read reuse}. 
    \item \textbf{Remove Attention-side $\mathcal{H}^{\mathrm{res}}$}: to enable exact dense-read reuse without a second full-state read, we remove residual mixing from the Attention sublayer, yielding
    \begin{equation}
    \label{eq:flash_attn_update}
    X_{\mathrm{act}}^{\mathrm{new}}
    =
    X_{\mathrm{act}}
    +
    p \odot
    \left(
    \mathcal{H}^{\mathrm{post,Attn}}
    \,\mathrm{out}_{\mathrm{Attn}}
    \right),
    \end{equation}
    where $\mathcal{H}^{\mathrm{post,Attn}}\in\mathbb{R}^{k\times 1}$ is the Attention post-mapping ($K_r{=}1$). This also removes a residual-mapping projection with a $kC\times k^2$ weight matrix and one Sinkhorn normalization per block; active-stream mixing is deferred to the MLP-side $\mathcal{H}^{\mathrm{res}}$.

    Since each active stream receives only a scaled copy of the Attention output, the change to the precomputed MLP base readout is a token-wise scalar multiple of $\mathrm{out}_{\mathrm{Attn}}$:
    \begin{equation}
    \label{eq:read_reuse}
    \mathrm{input}_{\mathrm{MLP}}
    =
    \mathrm{inp}_{\mathrm{M}}
    +
    \alpha\,\mathrm{out}_{\mathrm{Attn}},
    \qquad
    \alpha
    =
    \sum_{j=1}^{k}
    \mathcal{H}^{\mathrm{pre,MLP}}_{\mathcal{I}_j}
    \,p_j\,
    \mathcal{H}^{\mathrm{post,Attn}}_j .
    \end{equation}
    Here, $\alpha$ is assembled from quantities already computed during the Attention update. Appendix~\ref{app:xhc_flash_two_sub} provides the derivation.
\end{itemize}

Algorithm~\ref{alg:xhc_flash} summarizes how xHC-Flash shares full-state operations and updates active streams within one Transformer block.

\begin{algorithm}[t]
\caption{xHC-Flash: one Transformer block}\label{alg:xhc_flash}
\begin{algorithmic}[1]
\REQUIRE $X \in \mathbb{R}^{N \times C}$
\STATE
\STATE \textbf{/* Joint pre-forward (once per block) */}
\STATE $s,\, \mathcal{H}^{\mathrm{pre}}_{\mathrm{A}},\, \mathcal{H}^{\mathrm{pre}}_{\mathrm{M}} \leftarrow \mathrm{JointFullStateMap}(\mathrm{flatten}(X))$
\STATE $\mathcal{I}, p \leftarrow \mathrm{Route}(s; m,k)$; \quad $X_{\mathrm{act}} \leftarrow \mathrm{Gather}(X, \mathcal{I})$ \hfill $//$ $k$ active streams
\STATE $\mathrm{inp}_{\mathrm{A}},\, \mathrm{inp}_{\mathrm{M}} \leftarrow \mathrm{JointDenseRead}(X,\mathcal{H}^{\mathrm{pre}}_{\mathrm{A}},\mathcal{H}^{\mathrm{pre}}_{\mathrm{M}})$
\STATE
\STATE \textbf{/* Attention (no residual mixing) */}
\STATE $\mathcal{H}^{\mathrm{post}}_{\mathrm{A}} \leftarrow f_{\mathrm{post}}^{\mathrm{A}}(X_{\mathrm{act}})$ \hfill $// \;\mathbb{R}^{k \times 1}$
\STATE $\mathrm{out}_{\mathrm{A}} \leftarrow \mathrm{Attn}(\mathrm{inp}_{\mathrm{A}})$
\STATE $X_{\mathrm{act}} \leftarrow X_{\mathrm{act}} + p \odot (\mathcal{H}^{\mathrm{post}}_{\mathrm{A}}\,\mathrm{out}_{\mathrm{A}})$
\STATE $\alpha \leftarrow \sum_{j=1}^{k} \mathcal{H}^{\mathrm{pre}}_{\mathrm{M},\mathcal{I}_j}\,p_j\,\mathcal{H}^{\mathrm{post}}_{\mathrm{A},j}$
\STATE $\mathrm{input}_{\mathrm{M}} \leftarrow \mathrm{inp}_{\mathrm{M}} + \alpha\,\mathrm{out}_{\mathrm{A}}$ \hfill $//$ exact dense-read correction
\STATE
\STATE \textbf{/* MLP (with residual mixing + temporal aug) */}
\STATE $\mathcal{H}^{\mathrm{res}}, \mathcal{H}^{\mathrm{post}}_{\mathrm{M}} \leftarrow f_{\mathrm{res}}(X_{\mathrm{act}}), f_{\mathrm{post}}^{\mathrm{M}}(X_{\mathrm{act}})$ \hfill $// \;\mathbb{R}^{k \times k},\; \mathbb{R}^{k \times K_r}$
\STATE $\mathrm{out}_{\mathrm{M}} \leftarrow \mathrm{MLP}(\mathrm{input}_{\mathrm{M}})$
\STATE $X_{\mathrm{act}} \leftarrow \mathcal{H}^{\mathrm{res}} X_{\mathrm{act}} + p \odot (\mathcal{H}^{\mathrm{post}}_{\mathrm{M}}\,\mathrm{TempAug}(\mathrm{out}_{\mathrm{M}}))$
\STATE
\STATE $X \leftarrow \mathrm{Scatter}(X, \mathcal{I}, X_{\mathrm{act}})$
\RETURN $X$
\end{algorithmic}
\end{algorithm}

\textbf{Four-sublayer extension.}
The design extends naturally from two to four consecutive sublayers. In xHC-Flash-4sub, a single routing decision is shared across the two-block group, while four separate sublayer-specific pre-mappings, $\{\mathcal{H}^{\mathrm{pre},q}\}_{q=1}^{4}$, and their corresponding non-active base readouts are jointly formed from the group-entry state. Unlike the two-sublayer case, later corrections must account for both Attention and MLP write-backs; the latter contains $K_r{=}4$ temporal branches and cannot always be reduced to a scalar multiple of a single output vector. We therefore use the active-stream state already carried through the four sublayers to form later inputs, instead of maintaining a separate delta buffer. The active-state read is fused with active-side mapping computation, so the input correction does not introduce an additional active-state pass and retains the $2C$ read and $C$ write correction cost used in Table~\ref{tab:overhead_analysis}. Residual mixing is applied only at the final MLP, followed by a single scatter to the full state. Appendix~\ref{app:xhc_flash_four_sub} gives the equivalent correction form and derivation.

\textbf{Performance.}
Table~\ref{tab:xhc_flash_results} reports the accuracy-efficiency trade-off. xHC-Flash matches full xHC in validation loss (1.983) while reducing I/O from $73.5C$ to $51C$. xHC-Flash-4sub further reduces I/O to $40C$, comparable to the $34C$ of mHC at $N{=}4$, while maintaining a clear loss advantage over mHC (1.984 vs.\ 2.004). The small loss change suggests that sharing routing decisions and generating sublayer-specific pre-mappings from a short-window entry state does not significantly reduce the benefit of residual-stream expansion; Appendix~\ref{app:xhc_flash_tradeoffs} provides a detailed discussion.

\begin{table}[t]
    \centering
    \small
    \caption{xHC-Flash comparison on the 10B MoE model. I/O is the estimated memory traffic (read + write) amortized per sublayer from Table~\ref{tab:overhead_analysis}.}
    \label{tab:xhc_flash_results}
    \begin{tabular}{lcc}
        \toprule
        Method & Val. Loss $\downarrow$ & I/O / sublayer \\
        \midrule
        Vanilla baseline & 2.029 & $3C$ \\
        mHC ($N{=}4$)   & 2.004 & $\mathbf{34}C$ \\
        xHC (N=16, k=4)             & \textbf{1.983} & $73.5C$ \\
        xHC-Flash        & 1.983 & $51C$ \\
        xHC-Flash-4sub   & 1.984  & $40C$ \\
        \bottomrule
    \end{tabular}
\end{table}

\subsection{Infrastructure Design}
\label{sec:infrastructure}

A direct implementation of xHC consists of many small, largely memory-bound operators that repeatedly access the residual state, materialize normalized features and intermediate mappings, and incur kernel-launch overhead. Our implementation targets the default $N{=}16$ and $k{=}4$ configuration and organizes these operations into two stages: mapping generation and mapping application. Within each stage, operations sharing the same inputs are fused to reduce memory traffic and kernel launches. Residual states are stored in bfloat16, and projection operands use bfloat16, while normalization statistics, routing, mapping coefficients and Sinkhorn iterations use float32.

\textbf{Mapping generation.}
The first stage generates the router logits and the three xHC mappings, $\mathcal{H}^{\mathrm{pre}}$, $\mathcal{H}^{\mathrm{post}}$, and $\mathcal{H}^{\mathrm{res}}$. The router and pre-mapping project the same flattened state $x\in\mathbb{R}^{NC}$ but use different normalizations: LayerNorm for routing and RMSNorm for $\mathcal{H}^{\mathrm{pre}}$. We avoid materializing either normalized state by applying the corresponding corrections after projection. Let $\mu$ and $\nu$ denote the mean and second moment of $x$:
\begin{equation}
    Z^{\mathrm{r}}=\frac{xW^{\mathrm{r}}-\mu\,\mathbf{1}^{\top}W^{\mathrm{r}}}{\sqrt{\nu-\mu^2+\epsilon}},
    \qquad
    Z^{\mathrm{pre}}=\frac{xW^{\mathrm{pre}}}{\sqrt{\nu+\epsilon}}.
    \label{eq:fused_normalized_projection}
\end{equation}
We concatenate the two weight matrices along the output dimension and compute the raw projections with a single matrix multiplication:
\begin{equation}
    [\widetilde{Z}^{\mathrm{r}},\widetilde{Z}^{\mathrm{pre}}]
    =
    x[W^{\mathrm{r}},W^{\mathrm{pre}}].
\end{equation}
The statistics $\mu$ and $\nu$ are computed once and shared by both normalization branches. A fused Triton kernel applies the normalization corrections, generates $\mathcal{H}^{\mathrm{pre}}$, and performs fixed-2 plus routed-Top-2 selection. Its backward kernel jointly computes gradients for the normalization corrections and pre-mapping activation, and propagates routing gradients only through the two selected logits.

After routing, a specialized gather kernel extracts $X_{\mathrm{active}}\in\mathbb{R}^{k\times C}$ and computes its flattened-state RMS statistic, avoiding a separate statistics pass. Reusing this statistic, a single active-state projection jointly generates the post- and residual-mapping logits:
\begin{equation}
    [Z^{\mathrm{post}},Z^{\mathrm{res}}]
    =
    \mathrm{RMSNorm}\!\left(\operatorname{vec}(X_{\mathrm{active}})\right)
    [W^{\mathrm{post}},W^{\mathrm{res}}].
\end{equation}
where $\operatorname{vec}(\cdot)$ flattens the active streams into a row vector.
With $k{=}4$, Sinkhorn normalization operates on a $4\times4$ residual-mapping matrix per token. The backward kernel recomputes the intermediate Sinkhorn states rather than storing them during the forward pass.

\textbf{Mapping application.}
The generated mappings are applied by two fused kernels. Before the Transformer sublayer, the dense pre-mapping read and sparse residual mixing are computed jointly:
\begin{equation}
    \mathrm{input}
    =
    \sum_{i=1}^{N}\mathcal{H}^{\mathrm{pre}}_i x_i,
    \qquad
    X_{\mathrm{mixed}}
    =
    \mathcal{H}^{\mathrm{res}}X_{\mathrm{active}}.
\end{equation}
Without fusion, the dense read $\mathcal{H}^{\mathrm{pre}}X$ and active-stream mixing $\mathcal{H}^{\mathrm{res}}X_{\mathrm{active}}$ would be launched as separate memory-bound kernels. We evaluate both operations in one Triton kernel, reducing kernel-launch overhead. More importantly, the backward kernel accumulates the dense-read and sparse-mixing contributions directly into the gradient of the full residual state, avoiding a separate $kC$-dimensional active gradient and its subsequent scatter-add.

After the sublayer, a second kernel fuses the post-mapping application, routing-weight scaling, residual addition, and sparse scatter:
\begin{equation}
    X_{\mathcal I}^{\mathrm{new}}
    =
    X_{\mathrm{mixed}}
    +
    p\odot
    \left(
    \mathcal{H}^{\mathrm{post}}
    \mathrm{out}_{\mathrm{aug}}
    \right).
\end{equation}
Fusing post-mapping with residual addition avoids materializing the $kC$-dimensional write-back update, eliminating one $kC$ write and its subsequent $kC$ read before scatter. We specialize this for the Attention $K_r{=}1$ path and the MLP $K_r{=}4$ path with temporal feature augmentation. The backward kernels combine active-gradient gathering with gradients of the sublayer output, post-mapping coefficients, and routing weights, reducing both intermediate tensors and kernel launches.

\textbf{Temporal feature augmentation.}
For the MLP path, a specialized kernel jointly evaluates the three causal depthwise-convolution branches with kernel sizes $\{4,8,12\}$, reducing kernel launches and allowing overlapping temporal inputs to benefit from cache reuse. This corresponds to $C$ input reads and $3C$ output writes in our memory-access accounting. The subsequent $K_r{=}4$ post-mapping kernel consumes the original MLP output and the three convolutional outputs directly, avoiding construction of a concatenated $[S,B,4,C]$ tensor and eliminating an additional $4C$ read and $4C$ write. In backward, a fused input-gradient kernel accumulates contributions from all three branches, while dedicated kernels compute their weight gradients.

Together, these kernels remove normalized full-state intermediates, reuse statistics and fuse both forward and backward mapping application. They reduce implementation overhead beyond the algorithmic memory traffic reported in Table~\ref{tab:overhead_analysis}; the remaining full-state projection and dense pre-mapping read motivate the cross-sublayer amortization introduced by xHC-Flash in \S\ref{sec:xhc_flash}.

\textbf{xHC-Flash kernel fusion.}
For xHC-Flash, we extend the fusion pattern across the two shared sublayers. The router and two sublayer-specific pre-mapping projections are packed into one full-state projection, with normalization corrections, pre-mapping activations, and routed-Top-2 selection applied during the split-dimension reduction. A second full-state pass jointly computes the two dense readouts and gathers the active streams, reusing each loaded state element across these operations. The read-reuse correction is implemented as a small kernel that forms the scalar in Eq.~\ref{eq:read_reuse} and updates the MLP base readout. Unlike full xHC, where residual mixing is applied before each sublayer, xHC-Flash defers active-stream mixing to the MLP side. This allows the MLP-side kernel to fuse residual mixing, $K_r{=}4$ post-mapping, routing-weight scaling, and sparse scatter into a single active-state update; backward kernels follow the same fusion boundaries.

\textbf{End-to-end throughput.}
We measure wall-clock training throughput on the 18B MoE model without pipeline-communication overlap, to avoid communication-schedule confounders. Our reimplemented mHC ($N{=}4$) fused kernels add approximately 15\% training overhead over the baseline. This is higher than the 6.7\% reported in mHC~\cite{DS_mHC}, but the results are not directly comparable because the prior measurement may use a different scale, parallelism setup, or overlap schedule. xHC-Flash-4sub adds approximately 11\% overhead on top of mHC, mainly from the expanded $N{=}16$ full-state projection, the dual dense-read pass, and backward residual-stream operations. With pipeline-communication overlap, such as DualPipe~\cite{liu2024deepseek}, the effective end-to-end overhead can be further reduced.

For inference prefill at 2K tokens, mHC adds 11.4\% extra overhead over the vanilla baseline, while xHC-Flash-4sub adds 12.9\%. This corresponds to only a 1.3\% increase in total prefill latency over mHC, aligning with our observation that most of the additional training overhead of xHC-Flash-4sub comes from backward computation rather than the forward residual-stream path.

\vspace{-6pt}
\section{Conclusion}
\vspace{-6pt}

We presented xHC, a method for making residual-stream expansion in HC-family models effective and affordable beyond the common $N{=}4$ setting. We showed that directly scaling $N$ in mHC is limited by two bottlenecks: insufficient write-back information and the cubic cost of residual-mapping generation. xHC addresses these bottlenecks with temporal feature augmentation and a sparse residual-stream architecture that preserves dense read access while updating only $k$ active streams. Across language model pre-training experiments, xHC scales to $N{=}16$ with $k{=}4$, significantly outperforms mHC and the vanilla baseline on 18B and 28B MoE models, improves matched-loss compute efficiency, and remains effective under Muon. xHC-Flash and fused kernels further reduce the memory and implementation overhead needed for practical large-$N$ training. These results support expansion rate as a practical scaling dimension for HC-family models.

\newpage

\bibliography{reference}
\bibliographystyle{abbrv}

\newpage

\appendix

\section{Experimental Details and Hyperparameters}
\label{app:training_details}

\begin{table*}[t]
    \centering
    \caption{Model configurations and training hyperparameters for the main model scales. The 2.5B setting is used for $N$-sweep experiments, the 10B setting for ablations, and the 18B/28B settings for main results.}
    \label{tab:hyperparams}
    \resizebox{0.8\textwidth}{!}{
    \begin{tabular}{l | cccc}
        \toprule
        \textbf{Attribute} & \textbf{2.5B} & \textbf{10B} & \textbf{18B} & \textbf{28B} \\
        \midrule
        Total Parameters            & 2.5B  & 10B    & 18B   & 28B   \\
        Active Parameters           & 0.5B  & 1.4B    & 1.7B  & 2.7B  \\
        \midrule
        Layers                      & 15    & 15    & 28    & 32  \\
        Leading Dense Layers        & 1     & 1     & 1     & 1  \\
        MoE Layers                  & 14    & 14    & 27    & 31  \\
        Hidden Dimension            & 1024  & 2080  & 2112  & 2560  \\
        FFN Dimension (dense layer) & 3104  & 6144  & 6144  & 6144  \\
        Attention Heads             & 8     & 16    & 16    & 20  \\
        KV Groups (GQA)             & 4     & 8     & 8     & 10  \\
        Head Dimension              & 128   & 128   & 128   & 128  \\
        Routed Experts              & 144   & 144   & 144   & 144  \\
        Shared Experts              & 1     & 1     & 1     & 1  \\
        Active Experts (Top-$k$)    & 8     & 8     & 8     & 8  \\
        Expert FFN Dimension        & 320   & 704   & 672   & 768  \\
        Shared Expert Dimension     & 320   & 704   & 672   & 768  \\
        Normalization               & \multicolumn{4}{c}{RMSNorm ($\epsilon = 10^{-5}$)} \\
        Position Embedding          & \multicolumn{4}{c}{RoPE ($\theta = 50000$)} \\
        Activation                  & \multicolumn{4}{c}{SwiGLU} \\
        \midrule
        xHC Expansion Rate $N$           & 16   & 16   & 16   & 16   \\
        xHC Active Streams $k$           & 4    & 4    & 4    & 4    \\
        xHC Fixed Streams $m$            & 2    & 2    & 2    & 2    \\
        Temporal Conv Branches $r$       & 3    & 3    & 3    & 3    \\
        Conv Kernel Sizes                & \multicolumn{4}{c}{\{4, 8, 12\}} \\
        Gram--Schmidt Orthogonalization  & \multicolumn{4}{c}{\cmark} \\
        Temporal Augmentation Position   & \multicolumn{4}{c}{After MLP (MoE FFN) only} \\
        Stream Router Activation         & \multicolumn{4}{c}{Sigmoid} \\
        Sinkhorn Iterations              & \multicolumn{4}{c}{20} \\
        Gating Factor Init $\alpha$      & \multicolumn{4}{c}{0.01} \\
        \midrule
        Sequence Length             & \multicolumn{4}{c}{8192} \\
        Vocab Size                  & \multicolumn{4}{c}{152064} \\
        Tokenizer                   & \multicolumn{4}{c}{Qwen2} \\
        Global Batch Size           & 160   & 288   & --    & 768  \\
        Training Tokens             & --    & --    & --    & --   \\
        Warmup Steps                & 500   & 500   & 500   & 500   \\
        Optimizer                   & \multicolumn{4}{c}{AdamW ($\beta_1{=}0.9$, $\beta_2{=}0.95$, $\epsilon{=}10^{-15}$)} \\
        Base Learning Rate          & 6.95e-4 & 4.82e-4 & 3.97e-4 & 3.5e-4 \\
        LR Schedule                 & \multicolumn{4}{c}{WSD (warmup--stable--decay, exponential decay)} \\
        Min LR Ratio                & \multicolumn{4}{c}{0.1} \\
        Weight Decay                & \multicolumn{4}{c}{0.1} \\
        Gradient Clipping           & \multicolumn{4}{c}{1.0} \\
        \bottomrule
    \end{tabular}
    }
\end{table*}

\textbf{Model Architectures.}
All MoE backbones follow a DeepSeekMoE-style design: one leading dense layer followed by a stack of MoE layers. Each MoE layer contains routed experts and shared experts with top-8 sigmoid routing. We use grouped-query attention (GQA) with query-key LayerNorm (QK Norm). Detailed configurations are listed in Table~\ref{tab:hyperparams}.

\textbf{xHC Configuration.}
Unless otherwise specified, xHC uses $N{=}16$ total expansion streams with $k{=}4$ active streams per layer, of which $m{=}2$ are always-active fixed streams and $k{-}m{=}2$ are dynamically selected by the stream router. The router is a single linear projection applied to the LayerNorm-normalized, flattened $N{\times}C$ residual state, producing $N$ stream scores through a sigmoid activation. Temporal feature augmentation uses $r{=}3$ causal depthwise convolution branches with kernel sizes $\{4, 8, 12\}$, applied only after MLP (MoE FFN) layers, followed by Gram--Schmidt orthogonalization. Sinkhorn normalization for $\mathcal{H}_l^{\mathrm{res}}$ uses 20 iterations, and the gating factor $\alpha$ is initialized to 0.01. 
During xHC training, we observe that rare extreme activations can hinder the convergence of Sinkhorn normalization, leaving some row sums greater than one and amplifying the forward signal.
To improve stability, we apply row-sum clamping after Sinkhorn normalization by replacing each row $\mathcal{H}_{l,i:}^{\mathrm{res}}$ with $\mathcal{H}_{l,i:}^{\mathrm{res}} / \max(\sum_j \mathcal{H}_{l,ij}^{\mathrm{res}}, 1)$.
This operation only rescales rows whose sums exceed one, and we observe that it stabilizes training without degrading the performance.

\textbf{$N$-sweep Configuration.}
The $N$-sweep experiments in Section~\ref{sec:n_scaling} use the 2.5B setting in Table~\ref{tab:hyperparams}. Table~\ref{tab:n_sweep_config} lists the xHC active-stream configuration for each expansion rate. For mHC, we sweep dense expansion rates $N\in\{2,4,8,16\}$, where all $N$ streams participate in residual mixing and write-back. Other training settings, including tokenizer, context length, data mixture, optimizer, learning-rate schedule, and training-token count, are kept matched across compared runs.

\begin{table}[h]
    \centering
    \small
    \caption{$N$-sweep configuration for xHC.}
    \label{tab:n_sweep_config}
    \begin{tabular}{ccc}
        \toprule
        Expansion rate $N$ & Active streams $k$ & Fixed streams $m$ \\
        \midrule
        2  & 1 & 0 \\
        4  & 2 & 1 \\
        8  & 4 & 2 \\
        16 & 4 & 2 \\
        \bottomrule
    \end{tabular}
\end{table}

\textbf{Training Hyperparameters.}
At each model scale, xHC, mHC, and the vanilla residual baseline are trained with matched optimization recipes and data budgets, so that differences can be attributed to the residual-stream design rather than the training setup. We use AdamW with $\beta_1{=}0.9$, $\beta_2{=}0.95$, $\epsilon{=}10^{-15}$, weight decay 0.1, and gradient clipping at 1.0. The learning rate follows a warmup--stable--decay (WSD) schedule with 500 warmup steps, exponential decay, and a minimum-to-peak learning-rate ratio of 0.1. The base learning rate and global batch size are scaled with model size, as listed in Table~\ref{tab:hyperparams}.

\textbf{Final Stream Reduction.}
For both mHC and xHC, the multi-stream hidden state is reduced back to a single hidden vector before the language-model head. Specifically, before applying the final RMSNorm and unembedding layer, we sum the $N$ residual streams within each token and use the resulting $C$-dimensional representation for prediction.

\textbf{Muon Optimizer Configuration.}
For the experiments in Section~\ref{sec:muon}, we use a hybrid setup: Muon is applied to backbone 2D weight matrices (attention, MLP/MoE, and MoE router projections), while AdamW is used for other parameters (embeddings, normalization, and all xHC-specific parameters). Muon uses momentum $\beta{=}0.95$, 5 Newton--Schulz iterations, and a matched-AdamW-RMS target of 0.2. Other hyperparameters and the model architecture follow the 18B setting in Table~\ref{tab:hyperparams}.

\textbf{Evaluation Protocol.}
All benchmarks are evaluated under the modified OpenCompass framework~\citep{2023opencompass}. We use two evaluation paradigms: \textit{1) Perplexity-based evaluation.} The model scores each candidate answer by computing its conditional log-likelihood given the prompt and few-shot exemplars; the option with the highest likelihood is selected. This is used for multiple-choice benchmarks where answer options are fixed: MMLU (5-shot), MMLU-Redux (5-shot), CMMLU (5-shot), CEval (5-shot), ARC-Challenge (25-shot), C3 (3-shot). \textit{2) Generation-based evaluation.} The model generates a free-form response, which is then parsed and matched against the reference answer (exact match or rule-based extraction). This is used for: MMLU-Pro (5-shot), BBH (3-shot), CommonsenseQA (7-shot), GSM8K (4-shot), HumanEval (0-shot, pass@1), and LCBench (5-shot).

\section{Scaling Law Experiment Configuration}
\label{app:scaling_law}

We train four model sizes for each method (vanilla baseline, mHC, and xHC) under matched recipes, spanning training compute from approximately $1.7 \times 10^{19}$ to $4.0 \times 10^{20}$ FLOPs. Table~\ref{tab:scaling_law_config} lists the key model configurations used for scaling-law fitting. All models follow the same DeepSeekMoE-style template as Table~\ref{tab:hyperparams}, with 144 routed experts, top-8 routing, one shared expert, and one leading dense layer, while varying model width, depth, and expert FFN size. Scaling laws are fitted on the final language modeling (LM) loss.

\begin{table*}[t]
    \centering
    \caption{Scaling-law model configurations. All models use context length 8192, RoPE~\citep{RoPE} ($\theta{=}50000$), 144 routed experts, top-8 routing, 500 warmup steps, and the same data mixture.}
    \label{tab:scaling_law_config}
    \resizebox{\textwidth}{!}{
    \begin{tabular}{cc cccccc ccc}
        \toprule
        Active        & Total   & \multirow{2}{*}{Layers} & MoE & Hidden & FFN Dimension & Expert FFN & Q/KV & Training & Global Batch & Base Learning \\
        Parameters$^\dagger$ & Parameters & & Layers & Dimension & (dense layer) & Dimension & Heads & Budget & Size & Rate \\
        \midrule
        180M   & 1.08B  & 16 & 15 & 656  & 1920 & 224 & 8/4  & -- & 128 & 8.28e-4 \\
        340M  & 2.20B  & 15 & 14 & 1024 & 3104 & 320 & 8/4  & -- & 160 & 6.95e-4 \\
        460M  & 3.74B  & 23 & 22 & 1040 & 3072 & 352 & 8/4  & -- & 208 & 5.74e-4 \\
        1.10B & 9.46B  & 15 & 14 & 2080 & 6144 & 704 & 16/8 & -- & 288 & 4.82e-4 \\
        \bottomrule
    \end{tabular}
    }
    \vspace{0.25em}
    \begin{minipage}{0.98\textwidth}
    \footnotesize
    $^\dagger$Activated parameters per token, excluding embeddings.
    \end{minipage}
\end{table*}

We fit a shifted power law $\mathcal{L}(C) = A C^{-\alpha} + E$ to the final LM loss of each method, where $C$ denotes training FLOPs and $E = 0.72$ is an estimated irreducible loss. Following our fitting code, we perform linear regression on $\log_2(\mathcal{L} - E)$ against $\log_{10} C$ and convert the resulting fit back to the shifted power-law form above. The fitted parameters are reported in Table~\ref{tab:scaling_law_params}.

\begin{table}[h]
    \centering
    \small
    \caption{Fitted scaling-law parameters.}
    \label{tab:scaling_law_params}
    \begin{tabular}{lcc}
        \toprule
        \textbf{Method} & $A$ & $\alpha$ \\
        \midrule
        Baseline & 109.303 & 0.0936 \\
        mHC ($N{=}4$) & 99.139 & 0.0920 \\
        xHC ($N{=}16$, $k{=}4$) & 97.703 & 0.0919 \\
        \bottomrule
    \end{tabular}
\end{table}

\section{Training FLOPs and Parameter Overhead Analysis}
\label{app:flops_analysis}

We derive the per-layer parameter overhead introduced by xHC and mHC and report the resulting training FLOPs cost.

\subsection{Per-Layer Parameter Overhead}

Each Transformer layer contains two sublayers (attention and MLP). Let $C$ denote the hidden dimension, $N$ the expansion rate, $k$ the number of active streams, $K_r$ the temporal augmentation rank, and $\{\kappa_1, \dots, \kappa_r\}$ the depthwise convolution kernel sizes. Temporal augmentation is applied only after the MLP sublayer.

\textbf{xHC.}
Each sublayer adds a stream router ($N^2C$), a pre-mapping generator ($N^2C$), and a residual-mapping generator ($k^3C$). The post-mapping generator operates at rank $K_r$ for the MLP sublayer (with temporal augmentation) and rank 1 for the attention sublayer. Depthwise convolution weights are added only to the MLP sublayer. The total per-layer count is:
\begin{equation}
\label{eq:xhc_params}
P_{\text{xHC}} = \bigl(4N^2 + 2k^3 + k^2 + k^2 K_r + \textstyle\sum_{i=1}^{r} \kappa_i\bigr) \cdot C \,.
\end{equation}

\textbf{mHC.}
All $N$ streams are active ($k{=}N$), with no router, no temporal augmentation ($K_r{=}1$), and no convolution:
\begin{equation}
\label{eq:mhc_params}
P_{\text{mHC}} = (4N^2 + 2N^3) \cdot C \,.
\end{equation}
The $2N^3 C$ residual-mapping term is the source of cubic cost scaling.

\textbf{Comparison.}
With $K_r{=}4$ and kernel sizes $\{4,8,12\}$, xHC has the following per-layer overheads under representative active-stream budgets:
\begin{align*}
P_{\text{xHC},\,N=4,\,k=2} &= 124\,C, &
P_{\text{xHC},\,N=16,\,k=4} &= 1256\,C, \\
P_{\text{mHC},\,N=4} &= 192\,C, &
P_{\text{mHC},\,N=16} &= 9216\,C.
\end{align*}
xHC at $N{=}16,k{=}4$ uses \textbf{7.3$\times$ fewer} parameters per layer than dense mHC at the same expansion rate, while providing the same $N{=}16$ residual-memory capacity.

\subsection{Training FLOPs Overhead}

The parametric training FLOPs overhead per token introduced by the residual-stream module is:
\begin{equation}
\label{eq:hc_flops}
F_{\text{HC}} = 6 \cdot P_{\text{HC}} \cdot L\,,
\end{equation}
where $P_{\text{HC}}$ is instantiated as $P_{\text{xHC}}$ or $P_{\text{mHC}}$ depending on the method. The factor of 6 accounts for forward and backward passes through parametric projections. Non-parametric operations such as Sinkhorn normalization, residual mixing, LayerNorm on the multi-stream state, and Gram--Schmidt projections are analyzed separately in the memory-access and infrastructure discussions.

We compute the backbone training FLOPs per token $F_{\text{backbone}}$ using the standard $6\times$ approximation for linear layers, plus the sequence-dependent attention cost $6 S n_q d_h L$ (assuming causal attention with average context length $S/2$) and the LM-head projection $6VC$~\citep{kaplan2020scaling}. The overhead percentage is $\Delta\% = F_{\text{HC}} / F_{\text{backbone}} \times 100$. Table~\ref{tab:flops_verification} reports the overhead across methods and scales.

\begin{table}[h]
    \centering
    \small
    \caption{Parameter and training FLOPs overhead. $P$/layer is the per-layer HC parameter count; $\Delta P$\% is the parameter overhead relative to the vanilla backbone's activated parameters; $\Delta F$\% is the training FLOPs overhead.}
    \label{tab:flops_verification}
    \begin{tabular}{l ccc ccc}
        \toprule
        & \multicolumn{3}{c}{\textbf{18B} ($C{=}2112$, $L{=}28$)} & \multicolumn{3}{c}{\textbf{28B} ($C{=}2560$, $L{=}32$)} \\
        \cmidrule(lr){2-4} \cmidrule(lr){5-7}
        & $P$/layer & $\Delta P$\% & $\Delta F$\% & $P$/layer & $\Delta P$\% & $\Delta F$\% \\
        \midrule
        mHC ($N{=}4$)            & 405K ($192C$)    & 0.5\% & 0.7\% & 492K ($192C$)    & 0.6\% & 0.5\% \\
        mHC ($N{=}16$)           & 19.5M ($9216C$)  & 26.3\% & 18.9\% & 23.6M ($9216C$) & 30.2\% & 22.3\% \\
        xHC ($N{=}16$, $k{=}4$) & 2.65M ($1256C$)  & 3.5\% & 4.1\% & 3.22M ($1256C$)  & 4.1\% & 3.0\% \\
        \bottomrule
    \end{tabular}
\end{table}

xHC at $N{=}16$ incurs only modest additional training FLOPs across the main model scales, while dense mHC at the same expansion rate requires ${\sim}19\%$ overhead at 18B. For xHC, the overhead percentage decreases with model scale because $F_{\text{backbone}}$ grows with width while $P_{\text{xHC}}$ scales only linearly in $C$. The parameter overhead follows the same trend, confirming that the sparse architecture keeps the cost of large-$N$ expansion modest across scales.

\section{Additional Ablation Studies}
\label{app:additional_ablations}

We provide additional ablations on temporal feature augmentation, covering where augmentation is applied, whether Gram--Schmidt orthogonalization affects performance, and whether multi-scale temporal components are beneficial. Unless otherwise specified, experiments are conducted under the same 10B MoE setting as Section~\ref{sec:ablation}, and we report validation loss on the Pile test set.

\begin{table}[h]
    \centering
    \small
    \caption{Additional ablations on temporal feature augmentation. Lower validation loss is better.}
    \label{tab:additional_ablation}
    \begin{tabular}{l c}
        \toprule
        \textbf{Variant} & Val. Loss $\downarrow$ \\
        \midrule
        xHC default & 1.983 \\
        w/o Gram--Schmidt orthogonalization & 1.984 \\
        + Attention-side temporal augmentation & 1.985 \\
        \bottomrule
    \end{tabular}
\end{table}

\textbf{Attention-side Temporal Feature Augmentation.}
In xHC, temporal feature augmentation is applied only after MLP (MoE FFN) layers rather than after attention layers.
This design follows the different roles of the two sublayers: attention already performs content-dependent information mixing across tokens, whereas the MLP sublayer processes each token independently and therefore provides a cleaner place to inject lightweight local contextual features.
We also test adding the same temporal convolution after attention layers.
As shown in Table~\ref{tab:additional_ablation}, this variant gives a similar validation loss but is slightly worse than the default design by 0.002, indicating that attention-side temporal augmentation brings little additional benefit while introducing extra computation and implementation complexity.
We therefore use temporal augmentation only after MLP layers throughout the paper.

\textbf{Effect of Gram--Schmidt Orthogonalization.}
Section~\ref{sec:enrich} argues that convolutional components can retain strong components aligned with the original layer output.
Directly combining these correlated components may amplify the original write-back direction and make write-back scaling less predictable.
We therefore apply Gram--Schmidt orthogonalization to remove collinear components and keep the augmented write-back components distinct.
As shown in Table~\ref{tab:additional_ablation}, removing Gram--Schmidt orthogonalization gives nearly the same validation loss as the default xHC variant at the 10B ablation scale, indicating that this normalization does not remove useful signal.
However, at the 18B scale, the cosine similarity between convolutional branches and the main branch can exceed 0.7, and removing Gram--Schmidt orthogonalization leads to training instability.
We therefore retain Gram--Schmidt orthogonalization as a low-cost conditioning step that improves large-scale training stability while preserving validation performance.

\begin{table}[h]
    \centering
    \small
    \caption{Ablation on the number of temporal convolution branches. All variants use dense mHC at $N{=}16$ without sparse residual updates.}
    \label{tab:conv_branch_ablation}
    \begin{tabular}{l c c}
        \toprule
        \textbf{Variant} & \textbf{Conv Branches} & Val. Loss $\downarrow$ \\
        \midrule
        mHC & 0 & 1.998 \\
        mHC + Temp Aug (single-scale) & 1 & 1.989 \\
        mHC + Temp Aug (multi-scale) & 3 & 1.984 \\
        \bottomrule
    \end{tabular}
\end{table}

\textbf{Effect of multi-scale temporal components.}
We ablate the number of temporal convolution branches used in temporal feature augmentation.
This experiment is conducted on dense mHC at $N{=}16$, without the sparse residual-stream architecture, so that the effect of enriching the write-back signal can be isolated.
As shown in Table~\ref{tab:conv_branch_ablation}, adding one causal convolution branch improves validation loss from 1.998 to 1.989, confirming that local temporal information provides useful additional write-back signal.
Using three branches with different kernel sizes further improves the loss to 1.984.
This supports our design choice: multiple convolution branches provide write-back components over different temporal ranges, making the augmented signal more diverse and more effective for large-$N$ expansion.

\section{Design Analysis of xHC-Flash}
\label{app:xhc_flash_analysis}

xHC-Flash removes Attention-side residual mixing so that later sublayer inputs can be recovered exactly from precomputed base readouts and the sparse write-back updates, without rereading the full $N$-stream state. We first derive the correction for the two-sublayer variant and then generalize it to xHC-Flash-4sub.

\subsection{Two-Sublayer Dense-Read Reuse}
\label{app:xhc_flash_two_sub}

Let $X^{(0)}=(x_1^{(0)},\ldots,x_N^{(0)})^\top$ denote the block-entry state, and let $\mathcal I=(\mathcal I_1,\ldots,\mathcal I_k)$ denote the active-stream indices. The MLP pre-mapping is generated from $X^{(0)}$, and its base readout is precomputed as
\begin{equation}
\label{eq:app_mlp_base_readout}
\mathrm{inp}_{\mathrm M}
=
\sum_{i=1}^{N}
\mathcal{H}^{\mathrm{pre,MLP}}_i x_i^{(0)}.
\end{equation}
Because the Attention sublayer does not apply residual mixing, its write-back changes active stream $\mathcal I_j$ by
\begin{equation}
\label{eq:app_attn_writeback}
x_{\mathcal I_j}^{(\mathrm A)}
=
x_{\mathcal I_j}^{(0)}
+
p_j\mathcal{H}^{\mathrm{post,Attn}}_j
\mathrm{out}_{\mathrm{Attn}},
\qquad j=1,\ldots,k,
\end{equation}
while inactive streams remain unchanged. Applying the fixed MLP pre-mapping to the post-Attention state gives
\begin{align}
\mathrm{input}_{\mathrm{MLP}}
&=
\mathrm{inp}_{\mathrm M}
+
\sum_{j=1}^{k}
\mathcal{H}^{\mathrm{pre,MLP}}_{\mathcal I_j}
p_j
\mathcal{H}^{\mathrm{post,Attn}}_j
\mathrm{out}_{\mathrm{Attn}}
\\
&=
\mathrm{inp}_{\mathrm M}
+
\alpha\,\mathrm{out}_{\mathrm{Attn}},
\end{align}
where
\begin{equation}
\alpha
=
\sum_{j=1}^{k}
\mathcal{H}^{\mathrm{pre,MLP}}_{\mathcal I_j}
p_j
\mathcal{H}^{\mathrm{post,Attn}}_j .
\end{equation}
Thus, $\alpha$ is a token-wise scalar assembled from dynamically generated mapping coefficients, and Eq.~\ref{eq:read_reuse} requires only a scalar--vector multiply and an addition after the base readout has been computed.

\subsection{Four-Sublayer Extension}
\label{app:xhc_flash_four_sub}

xHC-Flash-4sub jointly forms four separate pre-mappings $\{\mathcal{H}^{\mathrm{pre},t}\}_{t=1}^{4}$ from the group-entry state, while sharing one routing decision $(\mathcal I,p)$ across the group. Unlike the two-sublayer case, later sublayers must consume both Attention and MLP write-backs; the MLP write-back has $K_r{=}4$ temporal feature components and therefore cannot always be reduced to a scalar multiple of a single output vector. We therefore express later inputs using a non-active base readout plus the current active-stream state, avoiding a separate accumulated-delta buffer.

For sublayer $q$, let $K_q$ denote the number of write-back components and let $\mathrm{out}^{(q)}_{\mathrm{aug},r}$ denote component $r$. Its sparse write-back to active stream $\mathcal I_j$ is
\begin{equation}
\label{eq:app_general_writeback}
\Delta x_{\mathcal I_j}^{(q)}
=
p_j
\sum_{r=1}^{K_q}
\mathcal{H}^{\mathrm{post},q}_{j,r}
\mathrm{out}^{(q)}_{\mathrm{aug},r}.
\end{equation}
In the two-sublayer xHC-Flash case, the only preceding write-back before the MLP input is the Attention write-back with $K_q{=}1$. The active-stream correction therefore factorizes into one scalar coefficient multiplying $\mathrm{out}_{\mathrm{Attn}}$, yielding Eq.~\ref{eq:read_reuse}. For any preceding write-back $q$, the active correction takes the form
\begin{equation}
\sum_{j=1}^{k}
\mathcal{H}^{\mathrm{pre},t}_{\mathcal I_j}
\Delta x_{\mathcal I_j}^{(q)}
=
\sum_{r=1}^{K_q}
\left(
\sum_{j=1}^{k}
\mathcal{H}^{\mathrm{pre},t}_{\mathcal I_j}
p_j
\mathcal{H}^{\mathrm{post},q}_{j,r}
\right)
\mathrm{out}^{(q)}_{\mathrm{aug},r}.
\end{equation}
When $q$ is an MLP sublayer, $K_q{=}K_r{=}4$, so this term is a weighted sum over multiple augmented output vectors rather than a scalar multiple of a single vector. A direct implementation could maintain these accumulated active-stream deltas explicitly, but doing so would introduce an extra $[S,B,k,C]$ buffer, including buffer updates after write-back and a later active-delta read for the correction.

Instead, for later sublayers we decompose the dense readout into a non-active base readout and the contribution of the current active streams:
\begin{equation}
\label{eq:app_four_sub_base}
\mathrm{base}_{\neg\mathcal I}^{(t)}
=
\sum_{i\notin\mathcal I}
\mathcal{H}^{\mathrm{pre},t}_i x_i^{(0)},
\qquad t=1,\ldots,4.
\end{equation}
Let $x_{\mathcal I_j}^{(t)}$ denote the current active-stream state before sublayer $t$. The input to sublayer $t$ is then
\begin{equation}
\label{eq:app_active_read}
\mathrm{inp}_{\mathrm{act}}^{(t)}
=
\sum_{j=1}^{k}
\mathcal{H}^{\mathrm{pre},t}_{\mathcal I_j}
x_{\mathcal I_j}^{(t)} .
\end{equation}
\begin{equation}
\label{eq:app_active_state_correction}
\mathrm{input}^{(t)}
=
\mathrm{base}_{\neg\mathcal I}^{(t)}
+
\mathrm{inp}_{\mathrm{act}}^{(t)} .
\end{equation}
Because residual mixing is deferred to the final MLP, intermediate sublayers update only the selected active streams, so
\begin{equation}
\label{eq:app_active_delta_identity}
x_{\mathcal I_j}^{(t)}
=
x_{\mathcal I_j}^{(0)}
+
\sum_{q<t}
\Delta x_{\mathcal I_j}^{(q)}.
\end{equation}
Substituting Eq.~\ref{eq:app_active_delta_identity} into Eq.~\ref{eq:app_active_read} decomposes the active contribution into its group-entry value and the accumulated sparse write-backs. Together with the non-active base in Eq.~\ref{eq:app_four_sub_base}, this recovers the same input that would be obtained by applying all preceding sparse write-backs to the group-entry readout, but without explicitly storing them.

\textbf{Implementation note.}
The derivation above uses the non-active-base form because it makes the input correction and memory accounting explicit: later sublayer inputs are formed from the unchanged non-active streams and the current active streams, without materializing a separate delta buffer. Our fused implementation follows the same correction mathematically, but may schedule the equivalent operations differently to improve kernel efficiency. These implementation choices do not change the resulting sublayer inputs.

\subsection{Exactness, Complexity, and Design Trade-offs}
\label{app:xhc_flash_tradeoffs}

This subsection clarifies what is exact in xHC-Flash and what is approximated relative to full xHC. The input corrections above are exact provided that (i) the routing decision and sublayer-specific pre-mappings are fixed from the sharing-window entry state, (ii) residual mixing is not applied before an intermediate sublayer, and (iii) each sparse write-back modifies only the selected active streams. Residual mixing at the final MLP does not violate these conditions because its input has already been formed and no subsequent sublayer in the sharing group requires another correction.

Under these conditions, the correction recovers the same sublayer input that would be obtained by applying the fixed pre-mapping to the updated residual state, without an $O(NC)$ dense read over all streams. For the two-sublayer case, this reduces to the scalar--vector form in Eq.~\ref{eq:read_reuse}. For xHC-Flash-4sub, the correction uses the active-state form in Eq.~\ref{eq:app_active_state_correction}, relying on the active streams already carried through the sharing group. If residual mixing were applied before an intermediate sublayer, the correction would additionally depend on the mixed active state and would no longer admit this simple active-state form.

Thus, the approximation in xHC-Flash is not the dense-read correction itself. Relative to full xHC, xHC-Flash approximates the dynamic stream-control schedule: routing is shared over the local window, sublayer-specific pre-mappings are generated from the window-entry state rather than from each updated intermediate state, and residual mixing is deferred to the final MLP in the sharing group. These choices are motivated by the sparse update structure: only $k$ of the $N$ streams change at each sublayer, while the exact correction exposes those changes to subsequent sublayers.

\textbf{Why the approximation is mild.}
The sharing window is short: one Transformer block for xHC-Flash and two blocks for xHC-Flash-4sub. Empirically, mHC at $N{=}4$ already achieves substantial gains with a much smaller stream-control space, suggesting that HC-family models do not require fully independent control at every sublayer to be effective. xHC-Flash also retains the main xHC mechanisms, including dense read access, separate sublayer-specific pre-mapping weights, sparse write-back to active streams, and MLP-side residual mixing. These factors help explain why the measured validation loss remains close to full xHC despite substantially lower memory traffic, with 1.983 for both full xHC and xHC-Flash, and 1.984 for xHC-Flash-4sub (Table~\ref{tab:xhc_flash_results}).

\end{document}